\documentclass[letterpaper]{article} 
\usepackage{aaai25}  
\usepackage{times}  
\usepackage{helvet}  
\usepackage{courier}  
\usepackage[hyphens]{url}  
\usepackage{graphicx} 
\usepackage{natbib}  
\usepackage{caption} 
\usepackage{algorithm}
\usepackage{algpseudocode}

\usepackage{bibentry}

\usepackage{amsfonts}
\usepackage{amsmath}
\usepackage{amsthm} 

\usepackage{graphicx}
\usepackage{subcaption}
\usepackage{caption}
\usepackage{array}
\usepackage{multirow}
\newtheorem{definition}{Definition}
\newtheorem{theorem}{Theorem}
\newcommand{\card}[1]{\left| #1 \right|}

\usepackage{newfloat}
\usepackage{listings}
\DeclareCaptionStyle{ruled}{labelfont=normalfont,labelsep=colon,strut=off} 
\floatstyle{ruled}
\newfloat{listing}{tb}{lst}{}
\floatname{listing}{Listing}
\usepackage{amsmath}
\nocopyright 

\title{Every Component Counts: Rethinking the Measure of Success for Medical Semantic Segmentation in Multi-Instance Segmentation Tasks}
\author {
   Alexander Jaus
   \textsuperscript{\rm 1,2},
   Constantin Seibold \textsuperscript{\rm 3,4},
   Simon Reiß \textsuperscript{\rm 1}, 
   Zdravko Marinov \textsuperscript{\rm 1,2}, \\
   Keyi Li\textsuperscript{\rm 1},
   Zeling Ye\textsuperscript{\rm 1},
   Stefan Krieg \textsuperscript{\rm 1},
   Jens Kleesiek \textsuperscript{\rm 4,5},
   Rainer Stiefelhagen \textsuperscript{ \rm1}
}
\affiliations {
\textsuperscript{\rm 1} Karlsruhe Institute of Technology, Karlsruhe, Germany \\
\textsuperscript{\rm 2} HIDSS4Health - Helmholtz Information and Data Science School for Health, Karlsruhe/Heidelberg, Germany \\
\textsuperscript{\rm 3} Institute for AI in Medicine, University Hospital Essen, Essen, Germany\\
\textsuperscript{\rm 4} Department of Nuclear Medicine, University of Duisburg-Essen\\
\textsuperscript{\rm 5} German Cancer Consortium (DKTK)- University Hospital Essen, Essen, Germany\\ 

}

\usepackage{bibentry}

\begin{document}

\maketitle

\begin{abstract}
We present Connected-Component~(CC)-Metrics, a novel semantic segmentation evaluation protocol, targeted to align existing semantic segmentation metrics to a multi-instance detection scenario in which each connected component matters. We motivate this setup in the common medical scenario of semantic metastases segmentation in a full-body PET/CT. We show how existing semantic segmentation metrics suffer from a bias towards larger connected components contradicting the clinical assessment of scans in which tumor size and clinical relevance are uncorrelated. To rebalance existing segmentation metrics, we propose to evaluate them on a per-component basis thus giving each tumor the same weight irrespective of its size. To match predictions to ground-truth segments, we employ a proximity-based matching criterion, evaluating common metrics locally at the component of interest. Using this approach, we break free of biases introduced by large metastasis for overlap-based metrics such as Dice or Surface Dice. CC-Metrics also improves distance-based metrics such as Hausdorff Distances which are uninformative for small changes that do not influence the maximum or $95^{\text{th}}$ percentile, and avoids pitfalls introduced by directly combining counting-based metrics with overlap-based metrics as it is done in Panoptic Quality. 
We will make the code for CC-Metrics publicly available.


\end{abstract}

\section{Introduction}

Semantic segmentation~\cite{long2015fully} is a cornerstone of medical image analysis as the automatic identification of critical areas such as organs-at-risk~\cite{lambert2020segthor} or metastases~\cite{gatidis2023autopet} might save valuable time in clinical care. With the ever-increasing performance of recent methods from the 3D-UNet~\cite{cciccek20163d}, transformer-based models~\cite{hatamizadeh2022unetr,hatamizadeh2021swin} to the nnUNet~\cite{isensee2021nnu}, segmentation seems to be on the cusp of clinical use. 
When trying to translate these algorithms to actual clinical use, however, these models with high dice scores tend to produce irresponsible errors such as the missing of novel smaller lesions which can significantly alter the treatment plan~\cite{haque2024development,de2024characterizing}. The question thus becomes, how could such issues be identified in the development process before stress testing on patients?

\begin{figure}
    \centering
    \includegraphics[width=0.8\linewidth, trim=0 220 460 0, clip]{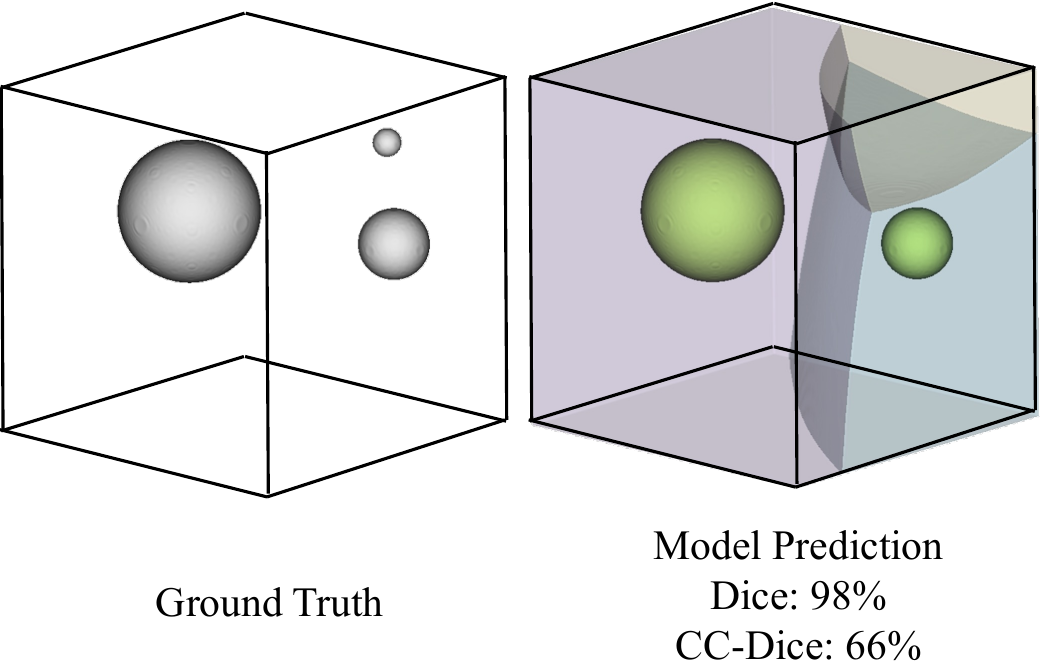}
    \caption{Reporting a Dice of 98\% in the shown example, highly overestimates the capability of the trained semantic segmentation model. This might leave radiologists with a false impression of how reliably the model can be used to predict tumors in body scans. 
    With CC-Metrics, we partition an image into distinct regions and evaluate standard semantic segmentation metrics on a per-component basis which gives each tumor the same importance.
    }
    \label{fig:title figure}
\end{figure}

In a typical cancer patient setup, we aim to identify models that can predict both large and small structures while also maximizing the overlap between the predicted tumor regions and the actual tumors.
This is non-trivial, as the selection of appropriate metrics for medical tasks depends on the specific scenario, the data at hand, the structure of outputs of the models and the type of questions the researcher tries to answer. Recent publications highlight the pitfalls of using suboptimal metrics~\cite{reinke2024understanding} and developed extensive recommendation frameworks~\cite{maier2024metrics}.

Despite the potential advantages of instance segmentation in distinguishing between overlapping objects, there seems to be less emphasis within the medical community on exploring and developing instance segmentation models for the purpose of volumetric multi-instance segmentation scenarios such as metastasis segmentation~\cite{gatidis2023autopet,oreiller2022head}.

We believe the main reason for this is: 
\textbf{Semantic Segmentation is sufficiently general}. Assume you have a perspective image of a street scene. Multiple objects and instances may overlap as the image is a projection of our three-dimensional world onto a two-dimensional plane. For instance, identifying individual people in a crowd does provide real value for downstream applications over a semantic-segmentation approach where a crowd would only be represented as a blob. This scenario, by definition, is not possible in volumetric images, each connected component is perfectly separated and there is no perspective overlap.
A consequence of this observation is that standard semantic segmentation is sufficiently general to solve detection as well as instance segmentation, by computing connected components in a post-processing step. We will refer to this setup as ``detection via segmentation''. 



Properly evaluating these semantic segmentation models in the context of a multi-instance scenario is challenging as semantic segmentation metrics are inherently not designed to care about instances and compute the agreement of predictions and ground truth globally. 

Within this work, we propose an embarrassingly simple, yet intuitive strategy for evaluating the performance of semantic segmentation models in the ``detection via segmentation'' setup by computing already established semantic segmentation metrics on a per-instance basis. By doing so, we give equal weight to each component reflecting their equal importance irrespective of their size. This approach aligns with the clinically motivated intent to treat all metastasis the same. 

To match predictions to ground truth values, we establish generalized Voronoi diagrams to partition each image into distinct regions, allowing predictions to be matched to the nearest ground truth connected component. By evaluating predictions locally, we eliminate the need for thresholds like those in Lesion Dice, allowing researchers to use existing, well-known metrics and avoiding the pitfalls of overlap-based matching or multiple true positives

Our contributions are as follows:
1) We formally derive CC-Metrics as a local, instance-aware metric evaluation protocol.
2) We analyze the shortcomings of using standard metrics in the ``detection via segmentation'' scenario and show how CC-Metrics deliver more informative insights.
3) We perform extensive simulations on PET/CT datasets showcasing the differences in evaluation styles under multiple scenarios.
4) We benchmark multiple segmentation models and compare their performance using standard and CC-Metrics.



\section{Related Work and Preliminaries}
Within this section, we briefly revisit common metrics used to evaluate semantic segmentation models and point out related work that has raised criticism regarding the presented metrics within the medical field.


\subsubsection{Overlap-based Semantic Segmentation Measures}
Overlap-based segmentation quality measures are one of the most frequently leveraged approaches to quantify the quality of segmentation masks. Typically the predictions $P$ are compared to the desired predictions $S$ by their area of overlap. Two of the most common metrics used to quantify this overlap are the Jaccard Index~\cite{jaccard1912distribution}, also called Intersection Over Union~(IoU), 

\begin{equation}
    \text{IoU} = \frac{|P \cap S|}{|P \cup S|}
\end{equation}

and the Dice coefficient~\cite{dice1945measures}, which is defined in terms of set cardinalities as

\begin{equation}
    \text{Dice} = \frac{2 \times |P \cap S|}{|P| + |S|}.
\end{equation}

A well-known limitation of these overlap-based metrics is their bias towards large objects and their inability to distinguish between different instances~\cite{reinke2024understanding}. The latest research thus recommends reporting counting-based metrics~(e.g., Precision) alongside overlap-based metrics~\cite{maier2024metrics}. To better counter size biases, a range of improvements, such as False Positive~(FP) and False Negative~(FN) penalties for the Dice score have been proposed~\cite{carass2020evaluating}.

\subsubsection{Unified-measures for Segmentation and Recognition Quality}


The Panoptic Quality~(PQ)~\cite{kirillov2019panoptic} metric, that was designed to unify detection and segmentation. PQ assigns predicted segments to ground truth segments by defining a match only if the predicted segment and ground truth segment overlap by at least $\text{IoU}>0.5$ rendering a guaranteed unique matching. As outlined in Eq.~\ref{Eq: PQ}, the metric is calculated by computing the average IoU of all True-Positives~(TP) and multiplying it by the F1-score~\cite{vanRijsbergen1979}. 


\begin{equation}
    \text{PQ}=\frac{\sum_{(p,g)\in \text{TP}} \text{IoU}(p,g)}{\card{\text{TP}}}\times \frac{\card{\text{TP}}}{\card{\text{TP}}+\frac{1}{2}\card{\text{FP}}+\frac{1}{2}\card{\text{FN}}}
    \label{Eq: PQ}
\end{equation}

The fixed-threshold-based matching approaches and direct metric combinations, like using the F1 score for counting and IoU for overlap, have limitations discussed in the \textit{Analysis of Segmentation Metrics} chapter. Other works have also criticized the usage of the PQ metric in cell nuclei segmentation~\cite{foucart2023panoptic}.

Recently, the Brats2023 challenge~\cite{moawad2023brain} started evaluating models based on a concept similar to Panoptic Quality by combining the overlap-based Dice score for all ground truth lesions, normalized by the number of all ground truth lesions and the number of FPs. They call this metric Lesion Dice (LD) and it is defined as follows:

\begin{equation}
    \text{LD}=\frac{\sum_{i\in (\text{TP}\cup \text{FN})}\text{Dice}(i)}{\card{\text{TP}}+\card{\text{FP}}+\card{\text{FN}}}
\end{equation}

The assessment of TP, FP, and FN is, as in PQ, based on an overlap-based criterion. However, LD does not demand the strict $\text{IoU}>0.5$ threshold which, as a consequence no longer guarantees a unique matching of prediction and ground truth.

\subsubsection{Boundary-based Semantic Segmentation measures}
\label{SS: RW Boundary-based Semantic Segmentation measures}
Boundary-based measures evaluate the quality of segmentations by focusing on the accuracy of the predicted boundaries relative to the ground truth. Two common metrics are Boundary IoU~\cite{cheng2021boundary} and (Normalized) Surface Dice~(NSD)~\cite{nikolov2021clinically,seidlitz2022robust}. These metrics modify the traditional overlap-based metrics by considering only pixels within a specified distance from the boundary, emphasizing edge alignment. They are particularly useful in applications where boundary accuracy is crucial, such as pathology delimitation. The NSD is defined as follows:

\begin{equation} \label{Eq:NSD}
    \text{NSD}=\frac{
    \card{S_\tau} + \card{P_\tau}
    }
    {
    \card{S} + \card{P}
    }
\end{equation}

where $S_\tau = \left\{ s \in S \mid \exists p \in P, \ d(s, p) \leq \tau \right\}$ is the set of surface pixels of the ground truth $S$ that are closer than a threshold $\tau$ to any of the surface pixels $p$ of the predicted surface $P$. $P_\tau$ and $P$ are defined similarly. For improved readability, we will use the term \textit{Surface Dice} interchangeably with\textit{ normalized Surface Dice}, with both terms consistently referring to the normalized metric throughout this work. 


\subsubsection{Distance-based Semantic Segmentation measures}

Distance-based measures quantify segmentation quality by evaluating the spatial distance between predicted and ground truth boundaries. The Hausdorff Distance~(HD)~\cite{hausdorff1914} measures the maximum distance from a point on the boundary of one set to the closest point on the boundary of another set, highlighting worst-case boundary errors. 
\begin{equation}
    \text{HD}= \max \left( \sup_{p \in P} \inf_{s \in S} d(p, s), \sup_{s \in S} \inf_{p \in P} d(s, p) \right)
\end{equation}
As this measure is susceptible to outliers, researchers have started to report the $95^{\text{th}}$ percentile, instead of the maximum distance which is known as the HD95 score. This worst-case behavior makes the metric less sensitive to small changes, especially for a large number of points~\cite{taha2015metrics}.



Another popular metric is the Average Surface Distance~\cite{heimann2009comparison} which computes the mean distance between corresponding points on the surfaces, providing an overall assessment of boundary alignment.

\section{CC-Metrics}
In this section, we introduce our proposed CC-Metrics evaluation protocol, focusing on the specific case of three dimensions ($d=3$) due to its relevance in medical volumetric multi-instance semantic segmentation. 

Consider an image $I\in \mathbb{R}^3$ and a binary target segmentation mask $S\in \{0,1\}^{h\times w \times d}$, where $h,w,d$ denote height, width and depth respectively. Further, consider a neural network $f$ computing binary predictions $P\in \{0,1\}^{h\times w \times d}$ given the image $f(I)=P$. To evaluate the quality $Q$ of the prediction $P$, given the target segmentation mask $S$, the standard approach is, to use a metric $m$, taking both $P$ and $S$ and computing $Q=m(P,S)$. Typically, metric $m$ generates a single quality measure, which globally measures some form of agreement between $P$ and $S$. In the default setup, all predictions $p\in P$ and target segmentations $s\in S$ will be passed to the metric which results in the default global quality measure $Q_{def}^{m}$. In the following, we derive CC-Metrics by first introducing a generalized Voronoi diagram, which partitions the image space according to connected components of the ground truth S.

\subsection{Definition of the Generalized Voronoi Diagram}
Consider the standard definition of a Voronoi diagram in a discrete three-dimensional metric space $\mathbb{T}$ with distance function $d$, where $\mathbb{T}$ is a set of discrete points such as the indices of a tensor, or the integer lattice points in $\mathbb{R}^3$. Let $K\subset \mathbb{T}$ be a set of indices~(3-tuples) which in our case can be thought of as a set of discrete locations in $\mathbb{T}$. To define the Voronoi diagram, a nonempty set of points $\{J_k\}_{k\in K}$ serves to define the sites of the diagram. Based on these definitions, a Voronoi Region $R_k$ is the set of all points in $\mathbb{T}$ that are closer to $\{J_k\}_{k\in K}$, measured by distance function $d$ than to any other site $\{J_l\}_{l\in K}$ with $k\neq l$. We define the distance function $d:\mathbb{T} \times K\mapsto \mathbb{R}: d(t,k)=\| t - k \|_2$ as a mapping from any location $t$ in $\mathbb{T}$ and the location of any point $\{J_k\}_{k\in K}$ to their Euclidean distance.
The standard Voronoi region $R_k$ is thus defined as 
\begin{definition}{Voronoi Region}
        \[
        R_k=\{t \in \mathbb{T}\ | d(t,J_k)< d(t,J_l)\quad \forall k,l \in K, k\neq l\}
        \]

    \label{Def: Voronoi diagram}
\end{definition}

This is the core concept according to which we plan to partition a given image tensor into regions. In our setup, the connected components of $S$ will serve as the sites according to which the Voronoi regions are defined. As these components in most cases contain more than one single n-tuple (location) per Vornoi site, we expand the previous Def.~\ref{Def: Voronoi diagram} by allowing connected components to form the sites. We expand the definition of $K$ by introducing $\mathbb{K}$ which bundles all points $\{J_k\}_{k\in K}$ into subsets based on their connectivity. 
Regarding the definition of connectivity, we rely on the concept of 26-connectivity. 
We define a function $\text{conn}_{26}:K\times K\mapsto \{\text{True},\text{False}\}$, which takes the locations of two points $J_{k_1}$ and $J_{k_2}$ as inputs and determines whether they are 26 connected. The points $J_{k_1}$ and $J_{k_2}$ are connected if $k_1=(a,b,c)$ and $k_2=(d,e,f)$ meet the following condition:
\begin{definition}{26-connectivity} \label{eq:26-connectivity}
    
\[
    \begin{aligned}
        &|a-d| \leq 1 \land |b-e| \leq 1 \land |c-f| \leq 1, \\
        &\text{with } (a, b, c) \neq (d, e, f)
    \end{aligned}
\]
\end{definition}

Leveraging the established Def.~\ref{eq:26-connectivity}, we can derive $\mathbb{K}$ which contains all sets of connected components according to the 26-connectivity criteria. 


\begin{definition}{Connected Component Sites}
\[
    \begin{aligned}[b]
    \mathbb{K} = & \{ C \subseteq K \mid \forall k_x, k_y \in C, \exists \mathrm{sequence} (k_{1}, k_{2}, \dots, k_{l}), \\
    & \text{with } k_x = k_1, k_y = k_l \text{ and } \mathrm{conn}_{26}(k_i, k_{i+1}) \, \forall i < l\}
    \end{aligned}
\]
\end{definition}

This updated definition of Voronoi sites which we now call $C$ reflects the generalized notion from single points to sets of connected points. We also modify the distance function to properly work on these sets of points, rather than two individual points. The updated function $d'$ should map a location $t\in \mathbb{T}$ and a set of points $\{J_C\}_{C\in\mathbb{K}}$ to a distance. The intuition is, that each location should still be matched to its closest site. We thus define $d':
\mathbb{T}\times \mathbb{K}\mapsto \mathbb{R}$ as $d'(t,C)=\min_{k\in C}\| t - k \|_2$ as the minimal distance to any location $k$ within the connecected component $C\in\mathbb{K}$.
With the updated, distance function, the generalized Voronoi region $R'$  can be easily defined as

\begin{definition}{Generalized Voronoi Region}
\[
\begin{aligned}[b]
R'_C = & \left\{ t \in \mathbb{T} \mid d'(t, J_{C_k}) < d'(t, J_{C_l}) \right. \\
      & \left. \forall C_k, C_l \in \mathbb{K}, \, C_k \neq C_l \right\}
\end{aligned}
\]
    \label{Def: Gen Voronoi diagram}
\end{definition}

\subsection{Calculating CC-Metrics}
After establishing the definition of the Generalized Voronoi regions, the calculations of CC-Metrics is straight forward. 

Let the current image $I$ be defined over a metric space $\mathbb{T}$ with indices $(a,b,c)$ for the 3D volume and $f(I)=P$ be the predictions of a neural network. The target segmentation mask $S\in\{0,1 \}^{h\times w \times d}$ can now be used to define the set of indices $K$ as 

\[
K = \{ (a, b, c) \in \mathbb{T} \mid S(i, j, k) = 1 \}
\]
In this definition, $K$ includes all the indices where the target segmentation mask has the value 1. Given $K$, $\mathbb{K}$ can be computed using the $\text{conn}_{26}$ function and $R'_C$ can be computed using Euclidean distance transforms. We show an algorithm for the computation of  $R'_C$ in the appendix.

We now define the local predictions $P_C$ for the region $R'_C$ as $P_C=P\cap R'_C$, where $P_C$ represents the set of predicted points in the Voronoi region $R'_C$. Similarly, we only consider the target segmentation within the same region $S_C=S\cap R'_C$. We now compute the metric of interest locally and separately for all regions

\[Q_C^m=m(P_C,S_C)\quad \forall C \in \mathbb{K} \]

ensuring that the evaluation is constrained to the specific region of interest.

We aggregate the different local quality measures using a standard average 

\[Q_{glob}^m=\frac{1}{\card{\mathbb{K}}}\sum_{C \in \mathbb{K}}Q^m_C\]

\section{Analysis of Segmentation metrics}
\label{S:Analysis}
Within the following section, we analyze failure cases of common semantic segmentation metrics used in a ``detection via segmentation'' scenario, based on a toy example. For all of the following analyses, we start, unless noted otherwise, with a ground truth consisting of three different sphere components. A visualization of this ground truth is given in Fig.~\ref{fig:title figure}. 
Initially, we assume a perfect prediction which we then degrade step by step as described in the following sections.

\subsection{Overlap Metrics: Dice vs. CC-Dice}
First, we compare the Dice metric as an example of an overlap-based metric. As an operation to degrade predictions, we continuously apply binary erosion to the prediction masks. We show the results of two experiments in Fig.~\ref{fig:Toy example dice erosion} in the top and middle plots. In the upper plot, we uniformly apply erosion to all components. It's important to note that binary erosion removes a larger percentage of volume from smaller spheres than from larger ones. As a result, we observe that the individual Dice scores of three components~(three black dashed lines) decrease with different speeds due to their different sizes. The global standard Dice coefficient (blue line) decreases more slowly and almost follows the Dice of the largest connected component. CC-Dice (orange line) decreases faster as it reflects the average per component Dice. 
In the second example (middle plot), we only apply erosion to the smallest component. The standard Dice is barely affected while CC-Dice quickly converges to $66\%$ reflecting perfect coverage for $2$ components and $0\%$ Dice for the smallest component.

\subsection{Unified Metrics: Panoptic Quality vs. CC-Dice}
\label{SS: Panoptic Quality vs CC-Dice}
Panoptic Quality~(PQ)~\cite{kirillov2019panoptic} is a metric that measures both recognition and segmentation quality, making it suitable for comparison with CC-Dice.

We compare CC-Dice and PQ in Fig.~\ref{fig:Toy example dice erosion} (top and middle plot) under the previously described scenarios. We observe two characteristic weaknesses of PQ. The first weakness is evident at erosion steps 3, 7, and 14 in the upper plot, where PQ drops rapidly. At other steps, the decline is smoother. This occurs because, at these erosion steps, the three different segmentation components fall below the $\text{IoU}>0.5$ thresholds relative to their ground truth, abruptly changing each component from TP to FP and causing sharp drops in the metric. In contrast, CC-Dice, which does not rely on fixed thresholds, behaves much more smoothly.

Another consequence of fixed thresholds is best observed in the middle plot of Fig.~\ref{fig:Toy example dice erosion}. Since PQ only measures segmentation quality for TPs, the score remains flat once the smallest component is no longer accepted as a TP. PQ does not reflect changes in segmentation quality after falling below this threshold. On the other hand, CC-Dice remains informative even for low overlaps.

A second major negative property of PQ is the direct combination of counting-based and overlap-based scores. As mentioned earlier, once the IoU of a component drops below 0.5, it is no longer considered a TP but becomes an FP, which negatively impacts the metric. However, after the component is completely eroded, the PQ score increases again because the presumed FP component is no longer present. This behavior is observed at erosion step 12 in the upper plot and erosion step 20 in the middle plot. We find this behavior suboptimal, as it introduces inconsistencies and irregularities in the evaluation of segmentation quality. 
The abrupt changes in PQ due to the fixed IoU threshold can complicate network evaluation, as minor variations in detected components may drastically alter the score. Additionally, the increase in PQ after a component is fully eroded is suboptimal, as it rewards the absence of predictions rather than partially correct ones. This is problematic in cases where missed components are more harmful than False Positives. While not designed to compete with PQ, CC-Dice provides a more consistent and representative assessment of segmentation quality by avoiding these issues.
\subsubsection{Lesion Dice vs. CC-Dice} Besides PQ we compare Lesion Dice (LD) as a second unified metric against CC-Dice.

LD requires the careful selection of many domain-specific hyperparameters, such as dilating the ground truth $n$-times to merge adjacent ground truth instances or ignoring predictions with components smaller than $k$ milliliters. Many of these parameters have to be chosen by experts~\cite{moawad2023brain}. While this adds flexibility, it places the burden of threshold selection on researchers and complicates cross-domain comparisons. In contrast, CC-Dice is a hyperparameter-free evaluation protocol.

Similar to PQ, LD directly incorporates the number of false positives into the score if they form components larger than $k$ milliliters. This creates a challenge in selecting an appropriate $k$ value. If $k$ is set high, the metric may ignore numerous false positives below $k$ without affecting the score. Conversely, if $k$ is low, even a few false positive pixels can significantly lower the score, in this setting, we penalize the model for uncertain predictions in a setting where the cost of a false positive is much lower than that of a false negative.

Unlike PQ, Lesion-Dice does not demand an $\text{IoU}>0.5$; even a single overlapping pixel can determine a match between prediction and ground truth. This no longer guarantees a unique matching, hence multiple masks with minimal overlap can be counted as TPs. 

\begin{figure}
    \centering
    \begin{subfigure}[b]{0.8\linewidth}
        \centering
        \includegraphics[width=\linewidth, trim=10 10 10 25, clip]{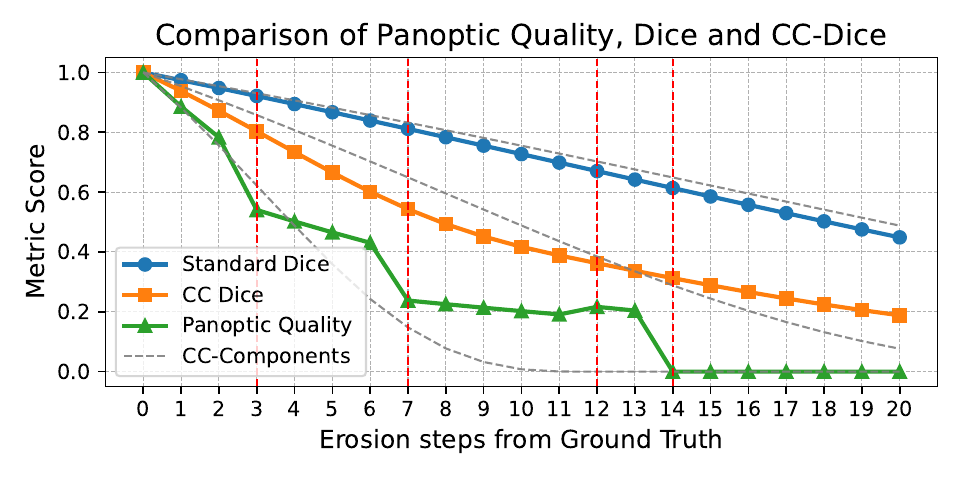}
    \end{subfigure}
    
    \vspace{0.2cm} 

    \begin{subfigure}[b]{0.8\linewidth}
        \centering
        \includegraphics[width=\linewidth, trim=10 10 10 25, clip]{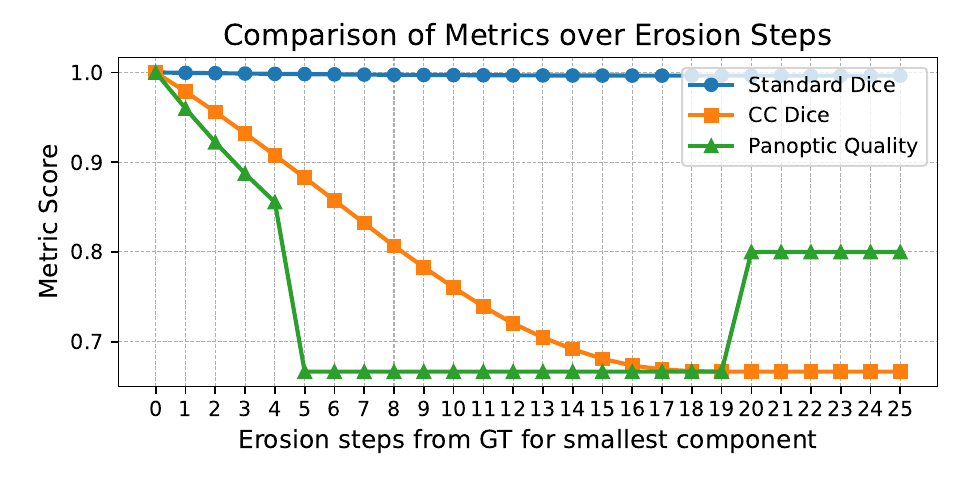}
    \end{subfigure}

\vspace{0.2cm}
    \begin{subfigure}[b]{0.8\linewidth}
    \centering
    \begin{subfigure}[b]{0.6\linewidth}
        \raggedright
        \includegraphics[width=1.1\linewidth, trim=10 10 10 10, clip]{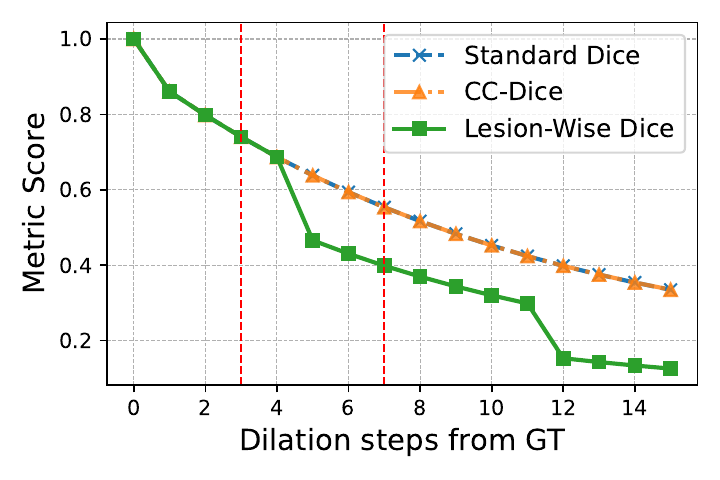}
    \end{subfigure}
    \begin{subfigure}[b]{0.38\linewidth}
        \raggedleft
        \includegraphics[width=0.8\linewidth, trim=0 265 770 0, clip]{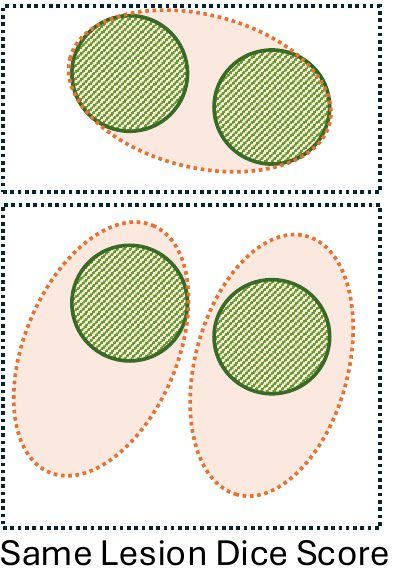}
    \end{subfigure}
\end{subfigure}

    \caption{Comparison of Dice, CC-Dice and Panoptic Quality: In the upper plot we start from a perfect prediction and degrade prediction quality by applying erosion to all components uniformly. In the middle plot we only degrade the prediction of the smallest mask. In the lower plot, we compare CC-Dice with Lesion Dice (LD) by using dilation to simulate oversegmentation (left) and highlight a pitfall of LD (right).}
    \label{fig:Toy example dice erosion}
\end{figure}

The lower plot of Fig.~\ref{fig:Toy example dice erosion} illustrates two pitfalls of LD. In the left graph, we start with three spheres of equal size and dilate the predicted masks. The Dice score (blue line) due to each component being of equal size behaves in line with CC-Dice (orange line). LD (green line) initially follows this pattern, however, once two previously not connected masks intersect, LS assigns the now connected mask as a TP to both lesions thereby decreasing the scores massively as seen at dilation steps 3 and 7. 

The double assignment in LD also leads to unexpected results, as demonstrated in the pitfall example on the right side of the lower plot. LD assigns the mask as a TP to both components individually. This leads to the scenario that two separate masks of the same size are not preferred over a single mask covering both components, despite the two masks covering many more false-positive locations. 



\begin{figure}
    \centering
    \begin{subfigure}[b]{0.8\linewidth}
        \includegraphics[width=\linewidth, trim=10 10 10 10, clip]{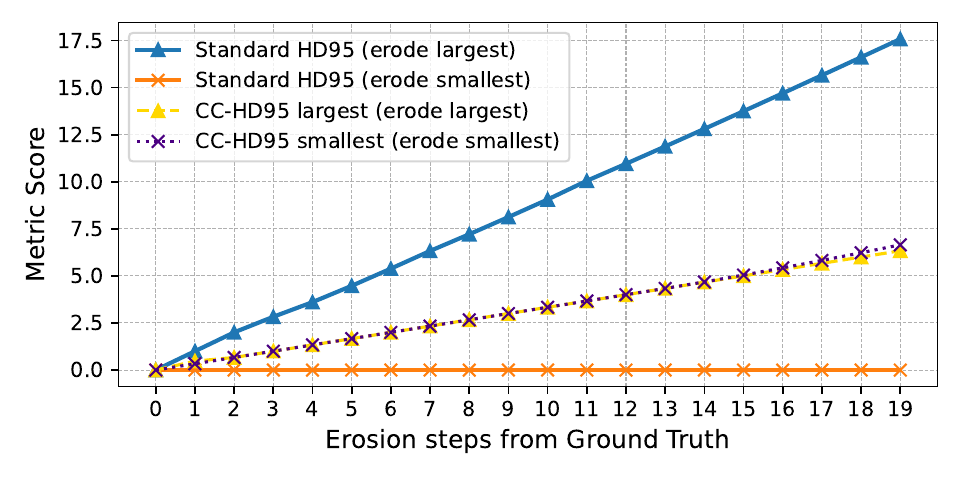}
    \end{subfigure}
    
    \vspace{0.2cm}
    
    \begin{subfigure}[b]{0.8\linewidth}
        \includegraphics[width=\linewidth, trim=10 10 10 25, clip]{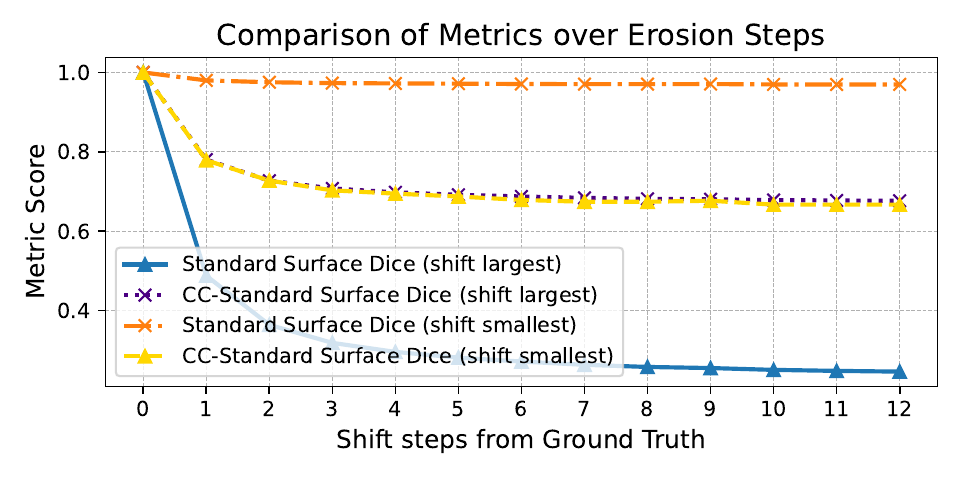}
    \end{subfigure}

    \caption{Comparison of the standard Hausdorff95 metric with the CC-Hausdorff95 metric (upper plot), as well as standard Surface Dice with CC-Surface Dice (lower plot). In both scenarios, we start from a perfect prediction and assess the metric scores while degrading the prediction quality of a large versus a small component.}
    \label{fig:Toy example hd erosion}
\end{figure}

\subsection{Distance Metrics: Hausdorff Distance vs. CC-Hausdorff Distance}

Distance-based metrics measure how far the boundaries of two masks are from each other. While this approach can be executed globally, large components, which naturally contain the vast majority of surface points, limit the importance of smaller components, either because their distances are treated as outliers and ignored as in Hausdorff95 or count little towards the average in Normalized Surface Distance. 
We explore this behavior in Fig.~\ref{fig:Toy example dice erosion}. Again we start with the spheres displayed in Fig.~\ref{fig:title figure} and assume a perfect prediction. We then gauge how the HD95 metric changes during the erosion of the smallest segmentation mask versus the erosion of the largest segmentation mask. It can be seen that using the standard evaluation protocol, the small component is ignored as its boundary distances are above the $95^{\text{th}}$ percentile, whereas eroding the largest component leads to a large increase in the metric. Evaluating the HD95 distance on a per-component basis normalizes the different number of boundary pixels and only compares one ground truth mask to its respective prediction. As desired, the CC-HD95 distances behave almost identically whether we erode the largest segmentation mask or the smallest segmentation mask.

\subsection{Boundary-based Metric: Surface Dice vs CC-Surface Dice}
Boundary-based metrics suffer from the same problem as overlap-based metrics when being evaluated globally. Large components having long boundaries limit the influence of smaller components. This behavior can be observed in the bottom plot of Fig.~\ref{fig:Toy example hd erosion}. In this example, we use a shift operation to degrade prediction quality, as erosion would leave the metric unchanged until the threshold $\tau$ in Eq.~\ref{Eq:NSD} is reached, at which point it would drop to zero. When degrading the largest component (blue line), the Surface Dice score decreases significantly, while degrading the smallest component (orange line) barely affects the overall score. Evaluating on a per-component basis results in consistent scores for the same operations, regardless of component size.

\section{Evaluation}
\label{S:Evaluation}
To evaluate the proposed CC-Metrics protocol on PET/CT datasets, we conduct experiments on two publicly available datasets:  AutoPET~\cite{gatidis2023autopet} and HECKTOR~\cite{oreiller2022head}, ensuring a comprehensive analysis across different cancer types. These datasets were selected as AutoPET involves patients with an average of 10 metastases, allowing us to assess performance in complex scenarios, while HECKTOR includes patients with an average of only 2 metastases per image, providing a setting where we expect standard metrics and CC-Metrics to be aligned.

We first simulate a range of different model failures using synthetic predictions. Given that HECKTOR includes only a few metastases per patient, AutoPET, with its higher average number of metastases, provides a more suitable dataset for our simulation. This choice allows us to effectively gauge how CC-Metrics behave in comparison to the standard evaluation protocol on a dataset level. Next, we train segmentation models on both the AutoPET and HECKTOR datasets and evaluate them using both approaches, ensuring that CC-Metrics and standard metrics are assessed across scenarios with varying metastasis counts.

\subsection{Evaluation on Synthetic Prediction}
\label{SS: Evaluation on Synthetic Predictions}
We initially assume a perfect prediction for each image and then apply a degrading function to progressively worsen and edit this prediction over $n$ steps. At each degrading step, we aggregate predictions, considering patients with at least $n$ metastases when altering $n$ components.

\begin{figure*}[ht]
    \centering
    \setlength{\tabcolsep}{2pt}
    \begin{tabular}{>{\centering\arraybackslash}m{0.01\textwidth} 
                    >{\centering\arraybackslash}m{0.18\textwidth} 
                    >{\centering\arraybackslash}m{0.18\textwidth} |
                    >{\centering\arraybackslash}m{0.18\textwidth} 
                    >{\centering\arraybackslash}m{0.18\textwidth} 
                    >{\centering\arraybackslash}m{0.18\textwidth}}
        \multicolumn{3}{c|}{\textbf{Drop n Components}} & \multicolumn{3}{c}{\textbf{Increase Nof. Oversegmented Components}}\\
        
        & 
        \small Dice \& CC-Dice & \small Surface Dice \& CC-Surface Dice & \small Dice \& CC-Dice & \small Surface Dice \& CC-Surface Dice & \small Hausdorff95 Distance  \& CC-HD95 Distance \\
        
        \rotatebox{90}{\shortstack{\small \textbf{Edit n smallest}}} &
        \includegraphics[width=0.18\textwidth, trim=30 10 40 30, clip]{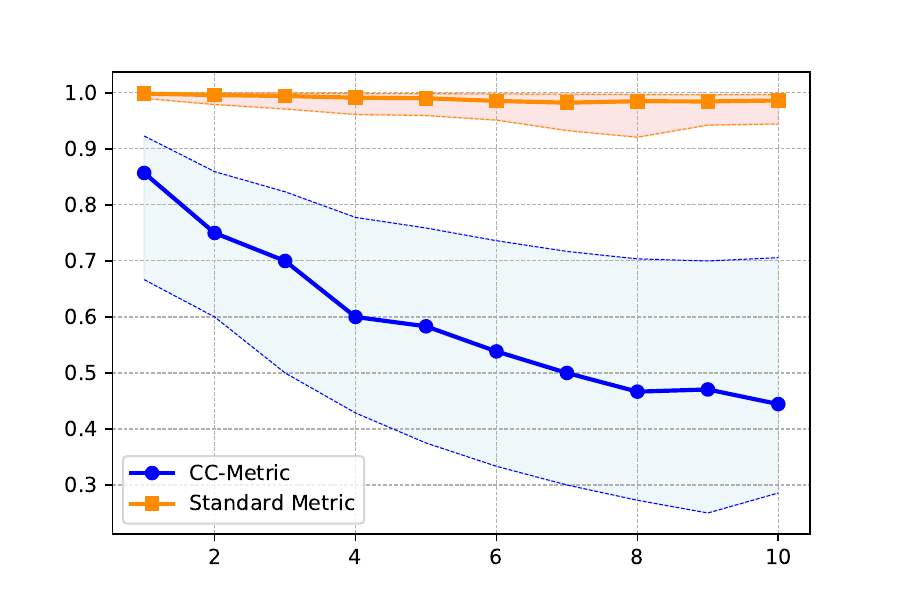} &
        \includegraphics[width=0.18\textwidth, trim=30 10 40 30, clip]{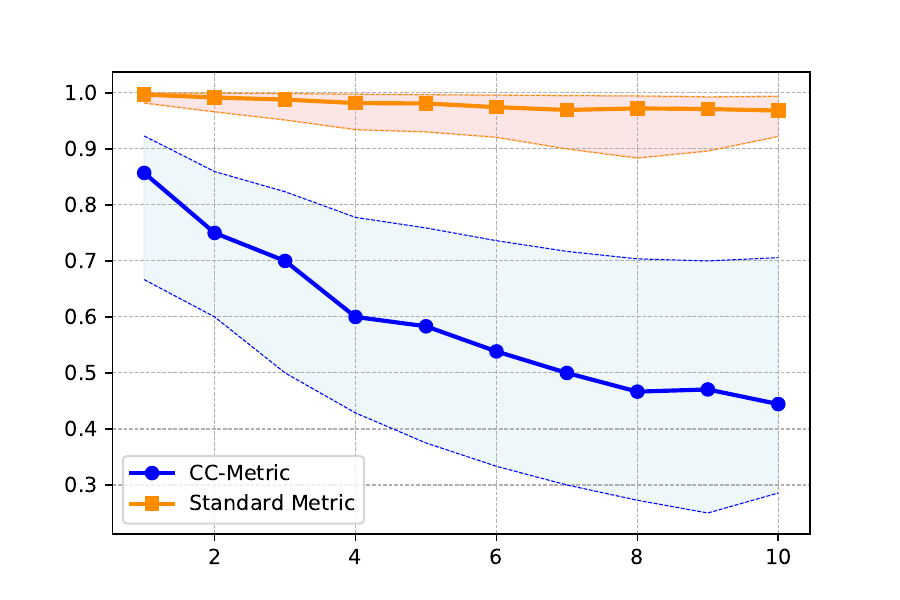} &
        \includegraphics[width=0.18\textwidth, trim=30 10 40 30, clip]{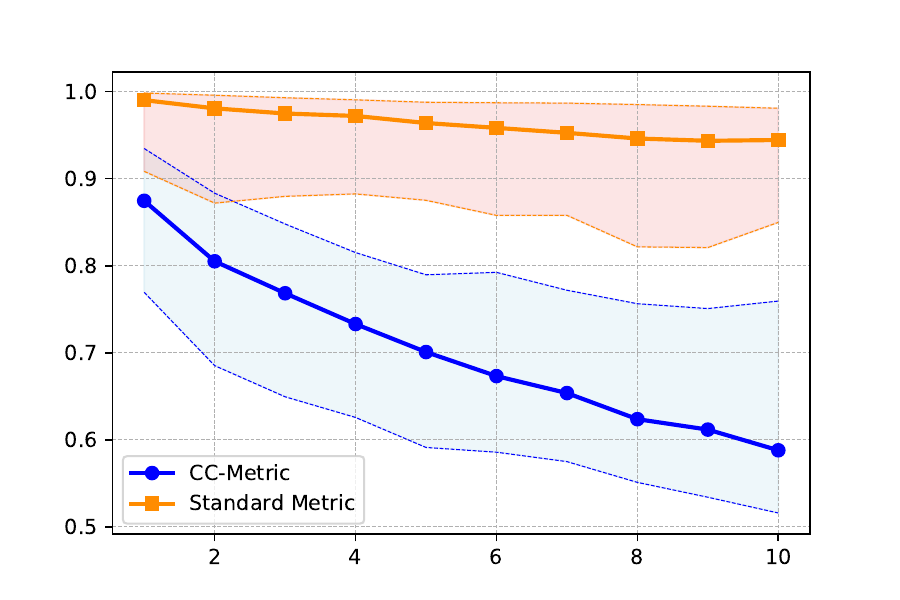} &
        \includegraphics[width=0.18\textwidth, trim=30 10 40 30, clip]{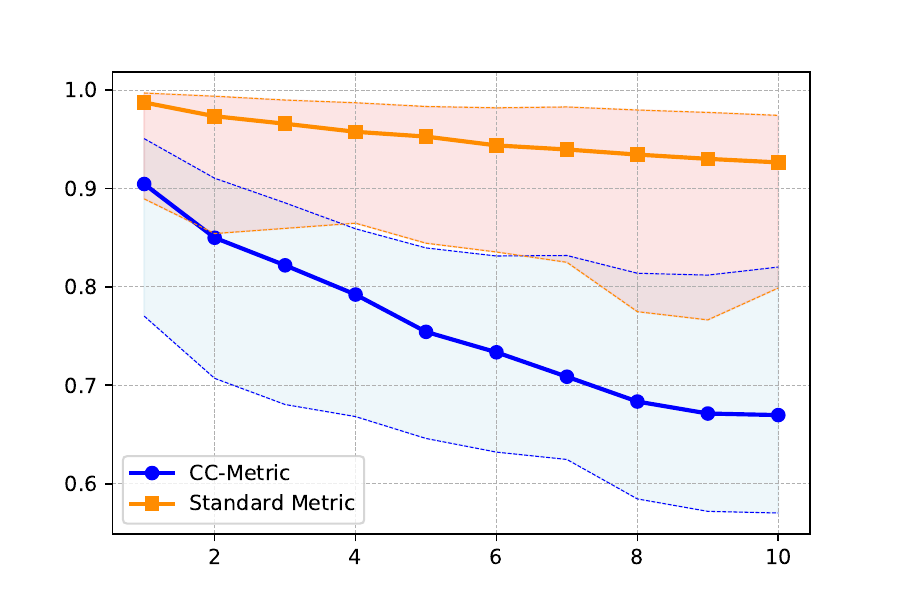} &
        \includegraphics[width=0.18\textwidth, trim=30 10 40 30, clip]{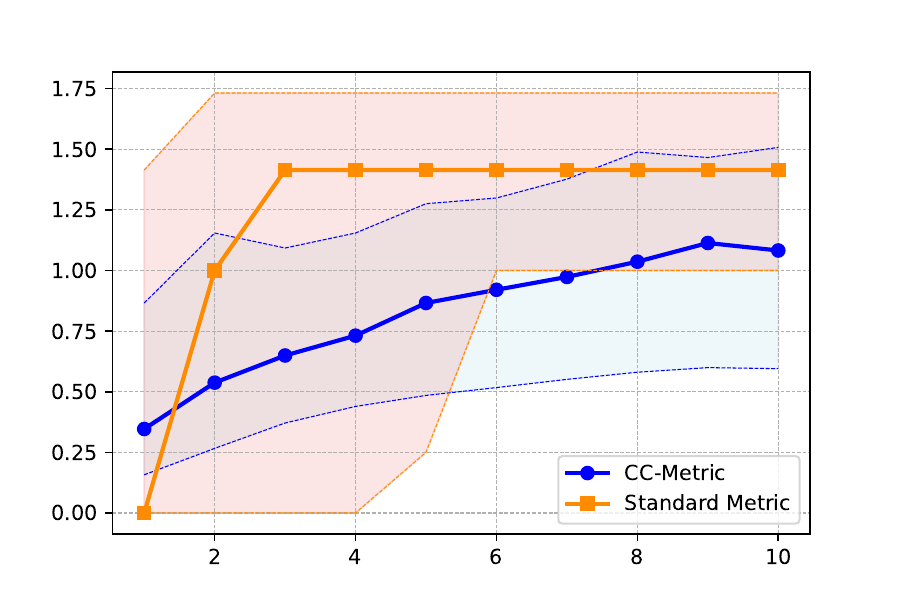} \\

        \rotatebox{90}{\shortstack{\small \textbf{Edit n largest}}} &
        \includegraphics[width=0.18\textwidth, trim=30 10 40 30, clip]{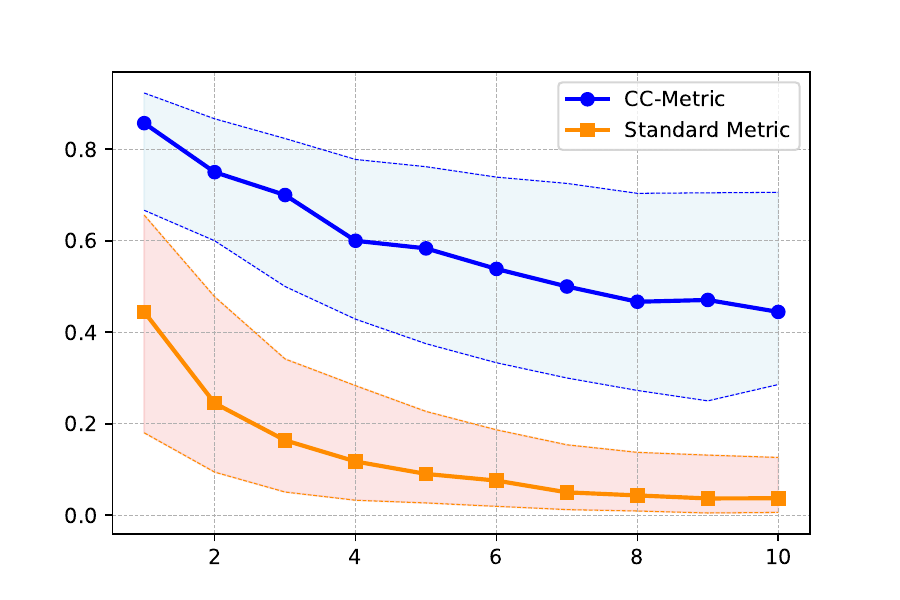} &
        \includegraphics[width=0.18\textwidth, trim=30 10 40 30, clip]{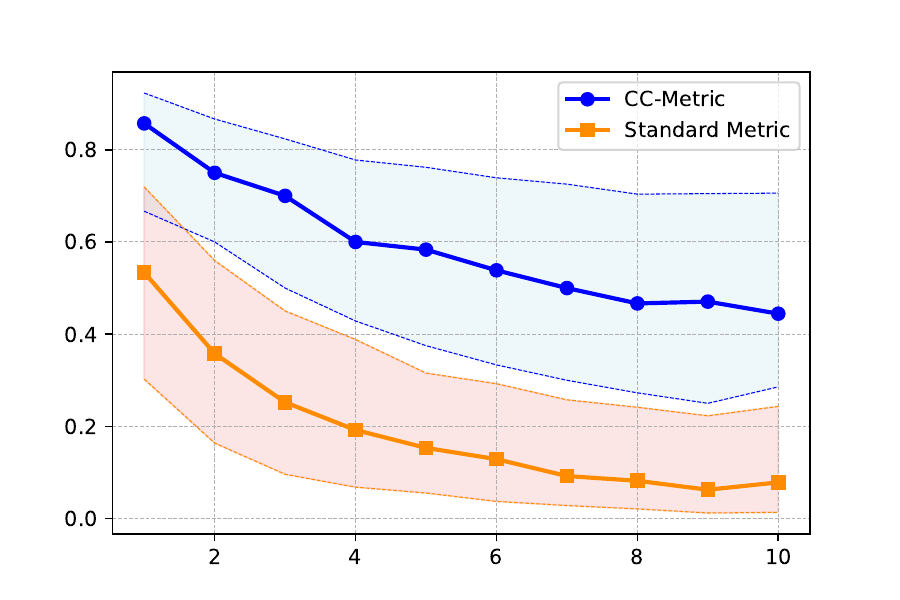} &
        \includegraphics[width=0.18\textwidth, trim=30 10 40 30, clip]{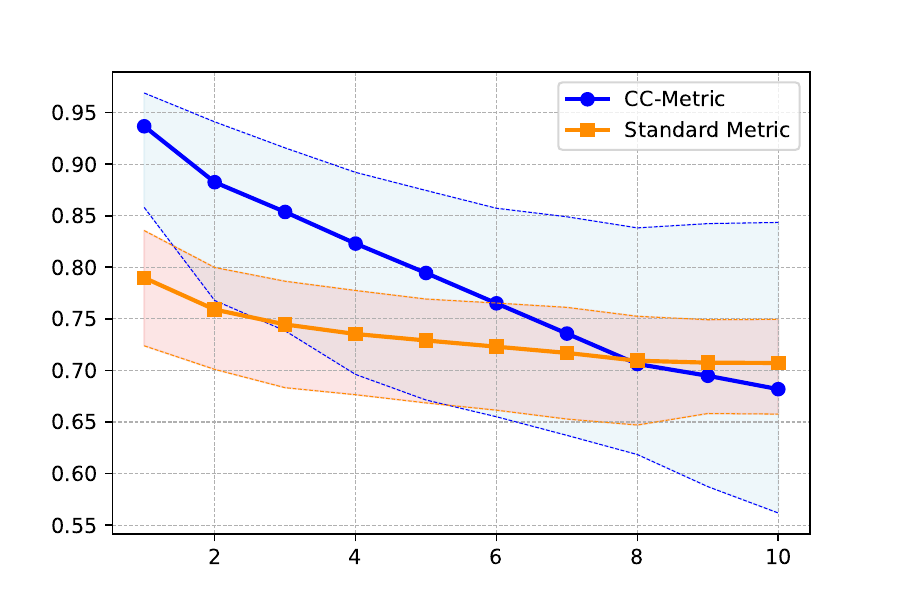} &
        \includegraphics[width=0.18\textwidth, trim=30 10 40 30, clip]{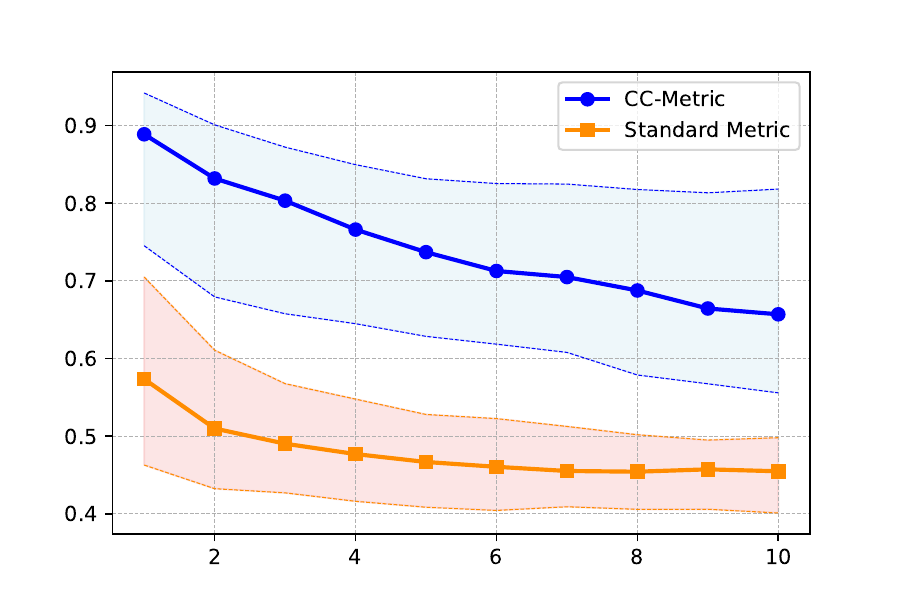} &
        \includegraphics[width=0.18\textwidth, trim=30 10 40 30, clip]{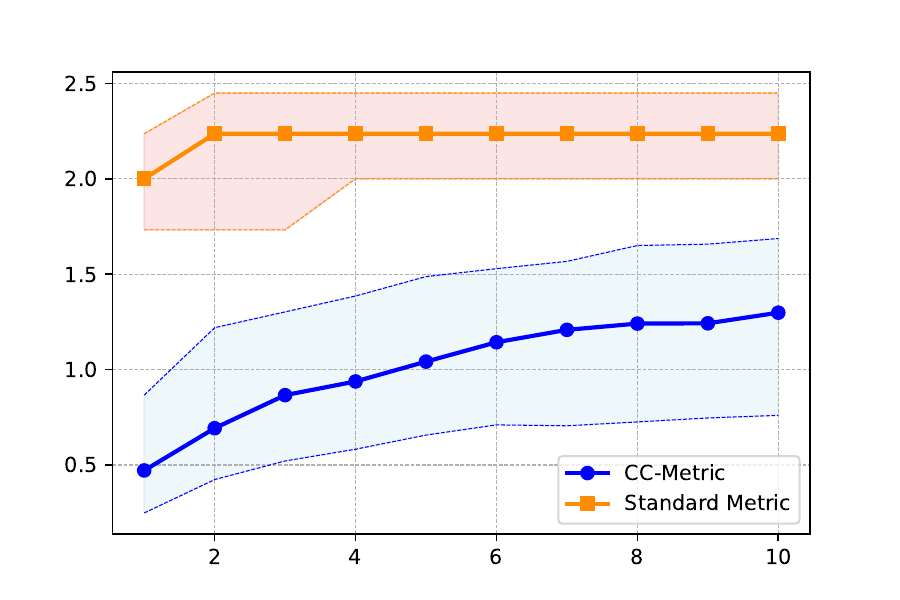} \\
    \end{tabular}


        \begin{tabular}{>{\centering\arraybackslash}m{0.01\textwidth} 
                    >{\centering\arraybackslash}m{0.18\textwidth} 
                    >{\centering\arraybackslash}m{0.18\textwidth} |
                    >{\centering\arraybackslash}m{0.18\textwidth} 
                    >{\centering\arraybackslash}m{0.18\textwidth} 
                    >{\centering\arraybackslash}m{0.18\textwidth}}
        \multicolumn{3}{c|}{\textbf{Insert n Random Components}} & \multicolumn{3}{c}{\textbf{Increase Nof. Undersegmented Components}}\\
        
        & 
        \small Dice \& CC-Dice & \small Surface Dice \& CC-Surface Dice & \small Dice \& CC-Dice & \small Surface Dice \& CC-Surface Dice & \small Hausdorff95 Distance  \& CC-HD95 Distance \\
        
        \rotatebox{90}{\shortstack{\small \textbf{Edit n smallest}}} &
        \includegraphics[width=0.18\textwidth, trim=30 10 40 30, clip]{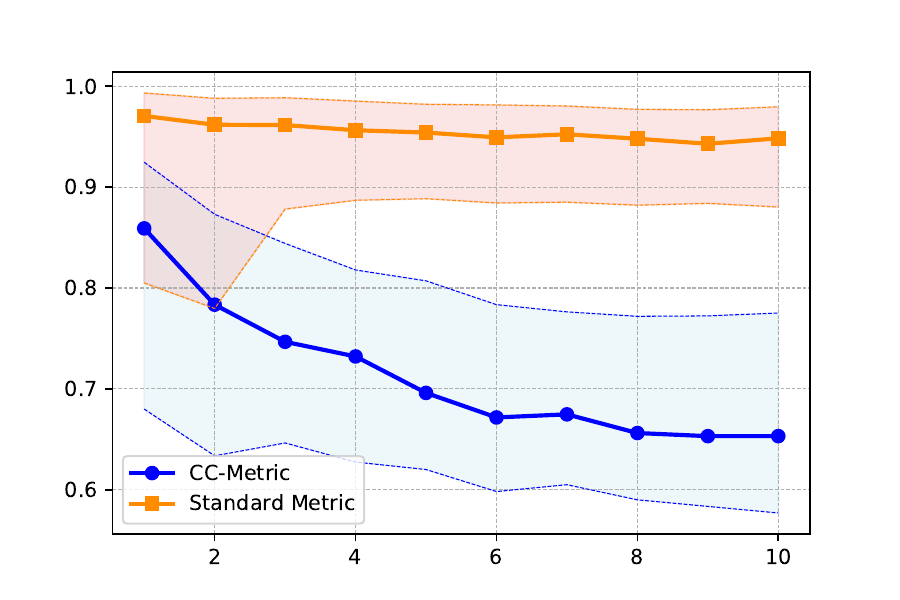} &
        \includegraphics[width=0.18\textwidth, trim=30 10 40 30, clip]{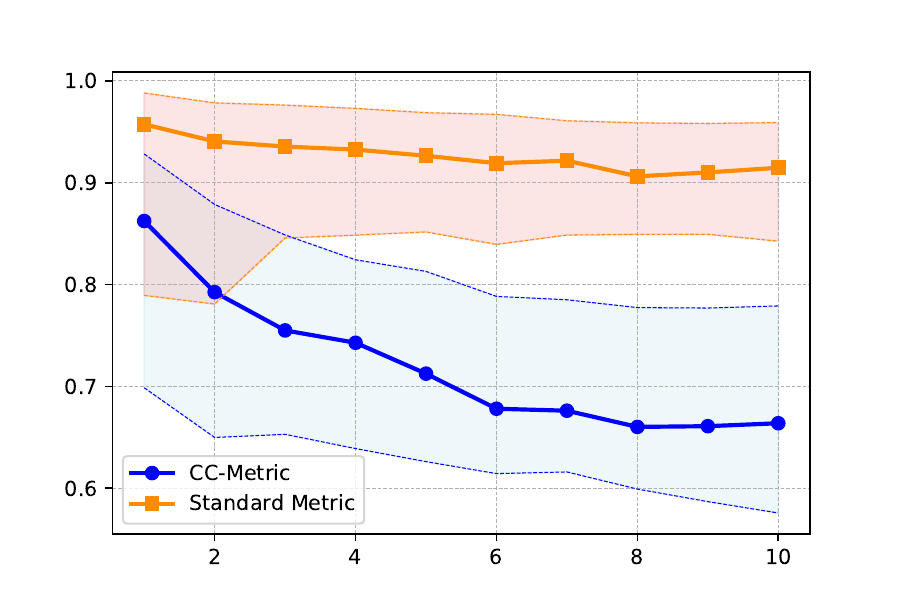} &
        \includegraphics[width=0.18\textwidth, trim=30 10 40 30, clip]{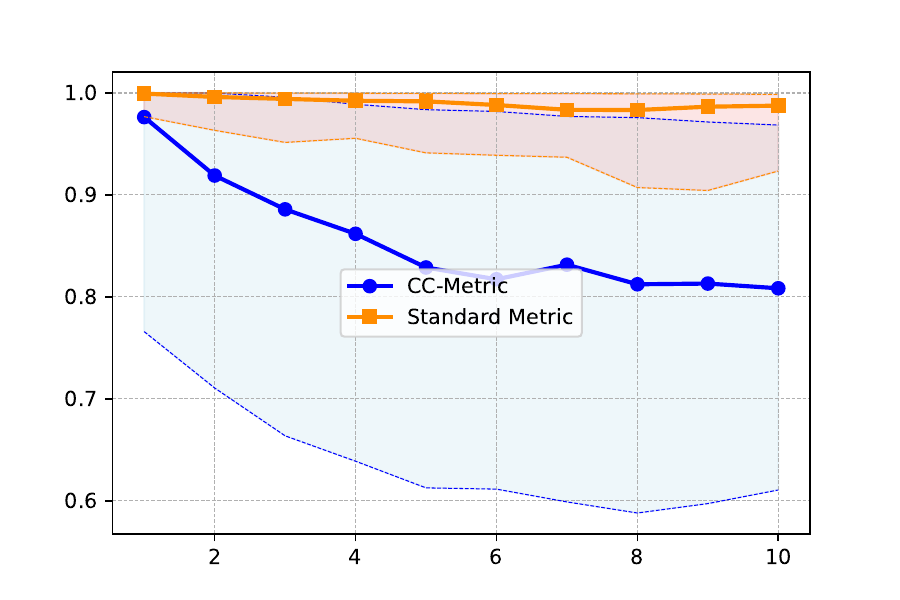} &
        \includegraphics[width=0.18\textwidth, trim=30 10 40 30, clip]{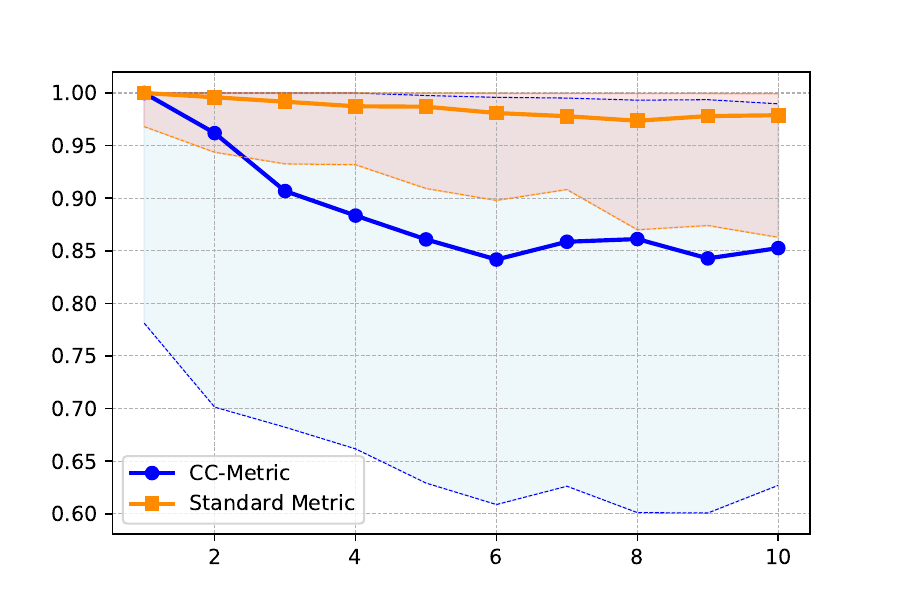} &
        \includegraphics[width=0.18\textwidth, trim=30 10 40 30, clip]{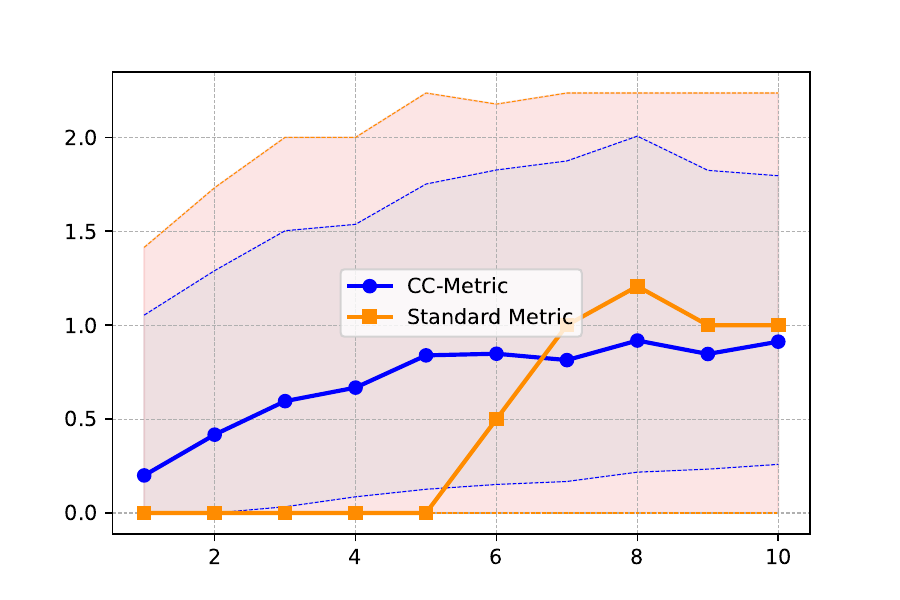} \\

        \rotatebox{90}{\shortstack{\small \textbf{Edit n largest}}} &
        \includegraphics[width=0.18\textwidth, trim=30 10 40 30, clip]{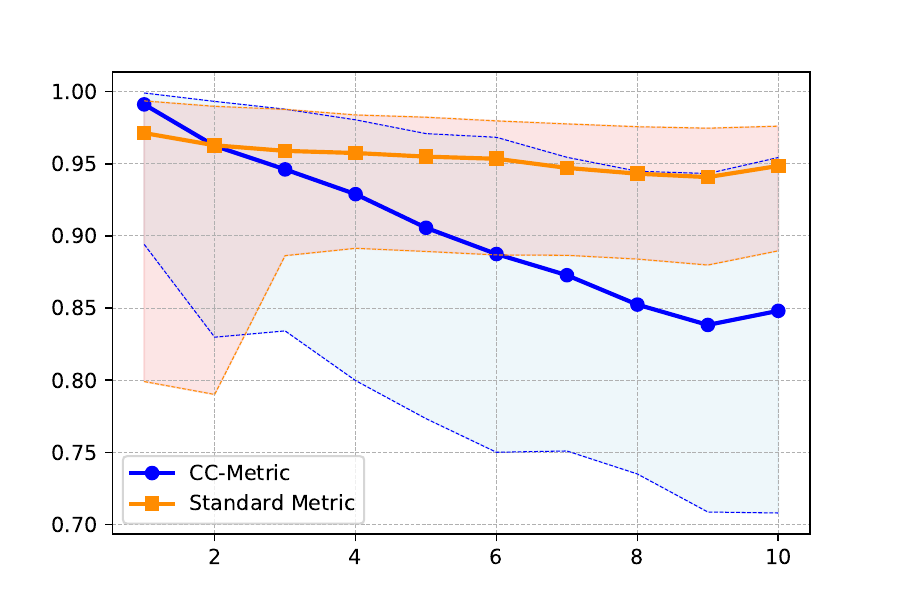} &
        \includegraphics[width=0.18\textwidth, trim=30 10 40 30, clip]{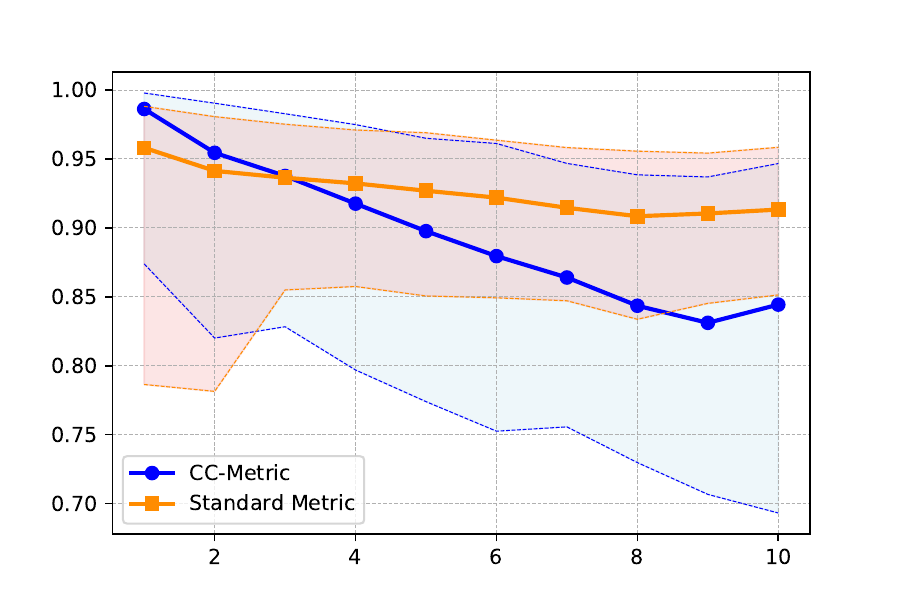} &
        \includegraphics[width=0.18\textwidth, trim=30 10 40 30, clip]{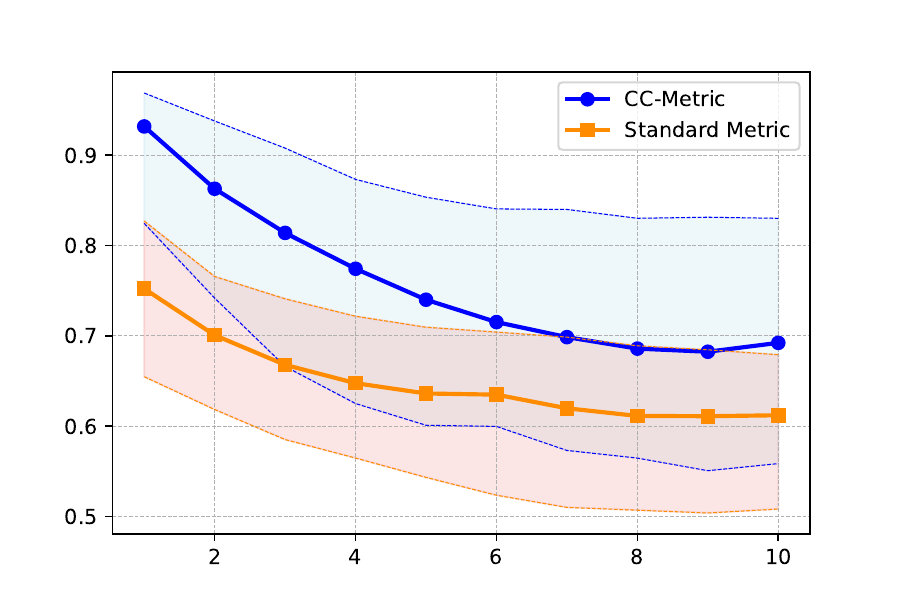} &
        \includegraphics[width=0.18\textwidth, trim=30 10 40 30, clip]{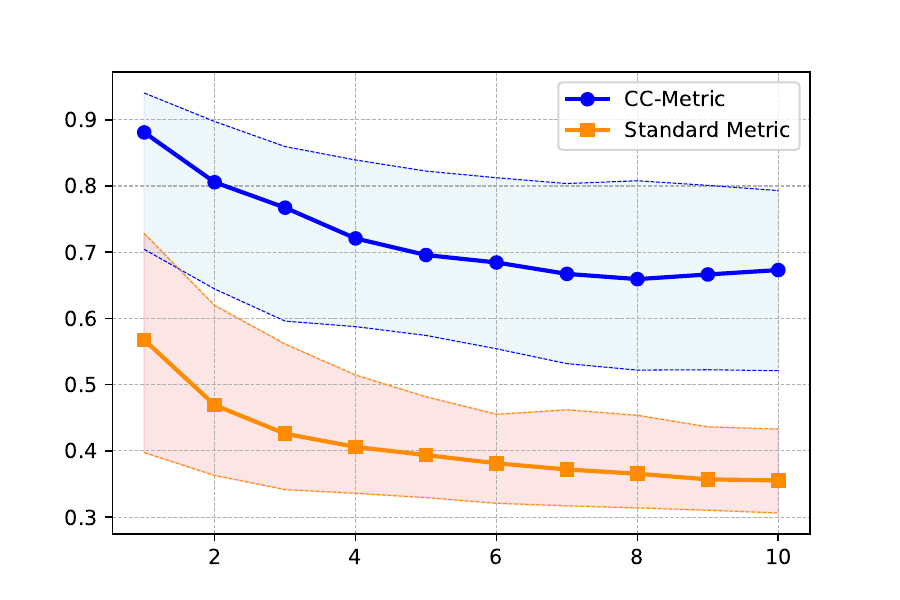} &
        \includegraphics[width=0.18\textwidth, trim=30 10 40 30, clip]{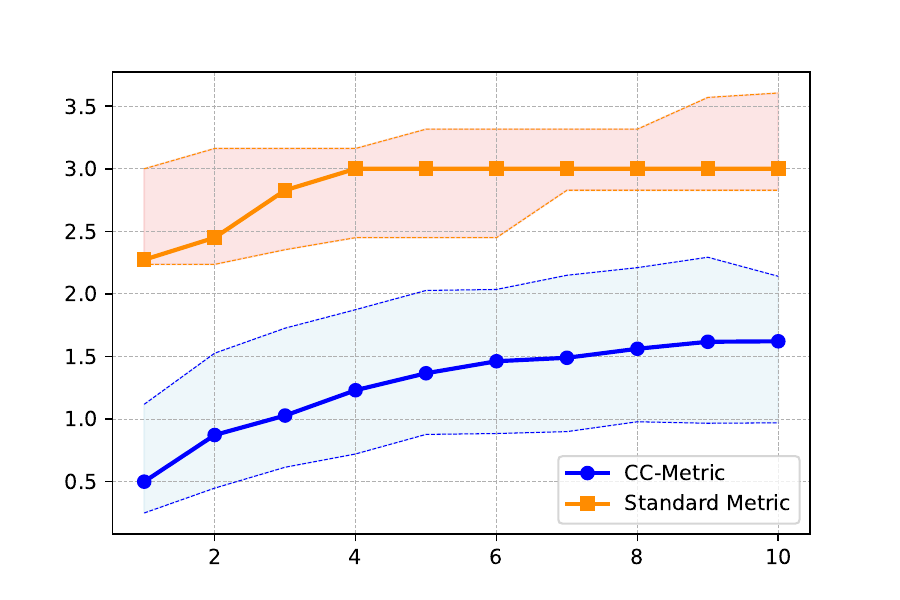} \\
    \end{tabular}
    
    \caption{Comparison of standard and CC semantic segmentation metrics on the AutoPET dataset across multiple scenarios}
    \label{fig:autopet sym results}
\end{figure*}

The simulation results for the AutoPET dataset are shown in Fig.~\ref{fig:autopet sym results}. The first row of each scenario simulates decreasing prediction quality for the $n$ smallest components, and the second row for the $n$ largest. We report the median along with the $25^{\text{th}}$ and $75^{\text{th}}$ percentiles of standard metrics in orange and CC-Metrics in blue.

\subsubsection{False Negatives:} We simulate the occurrence of false negatives by dropping components. We observe that both standard Dice and Surface Dice are heavily skewed towards large metastases. Even in the scenario where the $10$ smallest metastases are not captured, the standard metrics are barely affected. In contrast, CC-Dice and CC-Surface Dice more accurately represent the expected degradation of prediction quality and behave similarly regardless of whether False Negatives are larger or smaller components. 

\subsubsection{Oversegmentation:} We observe that overlap-based metrics as well as boundary-based metrics are dominated by large components. The Distance-based metrics HD95 and CC-HD95 have the capability to capture deviations of small and large components. A major downside of the standard Hausdorff metrics is, however, that they stay constant, reporting the global maximum distance for the standard HD and the global $95^{\text{th}}$ percentile for the HD95 metric. 
Due to this worst-case behavior of the metrics, a single component dominates the score, rendering how well other components are segmented whose scores fall below the global worst-case irrelevant.
CC-HD95 on the other hand reports the average per-component worst case and provides a more nuanced signal, allowing each component to influence the score. 

\subsubsection{Undersegmentation:}
In this scenario, we simulate under-segmentation where the model misses parts of the tumor or metastasis. We again observe CC-Metrics offering more nuanced insights into the prediction quality. Note that the standard HD95 metric first ignores errors when degrading the smallest errors as they fall below the $95^{\text{th}}$ percentile due to the large number of global boundary pixels. Again CC-HD95 is more informative.

\subsubsection{False Positives:}
In this scenario, we add one component to each of the $n$ selected regions. For each tumor volume, we randomly sample a location within the region defined by the $n$ largest or smallest components and insert a tumor there. The inserted tumors have a volume representing the 25th percentile of all metastasis volumes in the patient. We observe that CC-Dice and CC-Surface Dice are generally more sensitive to false positive predictions than standard Dice and Surface Dice under this simulation scenario.

\subsection{Evaluation of Real Predictions}
\label{SS: Evaluation on Real Predictions}
We train 4 segmentation models, namly nnUNet~\cite{isensee2021nnu}, DynUNet~\cite{isensee2019automated}, UNETR~\cite{hatamizadeh2022unetr}, and SwinUNETR~\cite{hatamizadeh2021swin}. 
Except for nnUNet, we use a consistent training process outlined in the appendix across all models with varying capabilities to assess how standard metrics and the novel CC-Metrics evaluate them, focusing on metric comparison rather than optimizing model performance.
We report the evaluation of the model predictions in Tab.~\ref{tab:model results}. 

\begin{table}[ht]
\caption{Comparing standard and CC Segmentation Metrics}
\centering
\small
\setlength{\tabcolsep}{5.5pt} 
\renewcommand{\arraystretch}{1.1} 
\begin{tabular}{@{}lcccccc@{}}
\hline
\multicolumn{7}{c}{\textbf{AutoPET: High Metastases Count Dataset}} \\ \hline
\textbf{Models} & \small Dice & \small CC-Dice & \small SD & \small CC-SD & \small HD & \small CC-HD \\ \hline
nnUNet & 67.5 & 47.4 & 66.3 & 52.5 & 71.1 & 46.4 \\
DynUnet & 33.5 & 30.5 & 28.5 & 32.8 & 190 & 114 \\
UNETR & 41.6 & 28.0 & 33.9 & 29.3 & 211 & 157 \\
SUNETR & 54.8 & 38.4 & 49.0 & 40.2 & 174 & 130 \\ \hline
\\
\hline
\multicolumn{7}{c}{\textbf{HECKTOR: Low Metastases Count Dataset}} \\ \hline
\textbf{Models} & \small Dice & \small CC-Dice & \small NSD & \small CC-SD & \small HD & \small CC-HD \\ \hline
DynUnet & 78.7 & 78.7 & 80.3 & 80.3 & 37.7 & 37.6 \\
UNETR & 60.8 & 60.7 & 56.8 & 56.8 & 151 & 151 \\
SUNETR & 72.3 & 72.3 & 71.2 & 71.3 & 113 & 113 \\ \hline
\end{tabular}
\label{tab:model results}
\end{table}

On the AutoPET dataset with its high number of average metastases per patient, we find significant differences between standard and CC-Metrics. The CC-Dice scores are significantly lower for all models than the traditional Dice scores. This disparity is most notable in models like nnUNet, where the Dice score is 67.5, but the CC-Dice score drops to 47.4, indicating that while traditional metrics might suggest a model performs well, the large difference between Dice and CC-Dice highlights the model is struggling with smaller metastases. We also find that ranking models by Dice and CC-Dice yields different results, with DynUNet outperforming UNETR in CC-Dice despite its significantly worse performance measured in standard Dice. Regarding the Surface Dice~(SD) score, we observe a similar picture as for the Dice. 
The Hausdorff95 distance, abbreviated as HD in the table, and its CC variant~(CC-HD) show marked differences, with CC-HD scores being generally lower, indicating better performance than the standard HD95. This result is to be expected as the standard metric focuses on global worst-case scenarios. The difference can be interpreted as the difference between the global worst-case and the average per-component worst-case scenario for the dataset. This effect is best shown in the simulation results in Fig.~\ref{fig:autopet sym results}.

On the HECKTOR dataset, we find CC-Metrics to be very well aligned with standard metrics for all models due to the low average number of ground truth components, highlighting that CC-Metrics do not bias results in unexpected ways.

\section{Discussion and Conclusion}

Within this work, we introduce CC-Metrics, an evaluation protocol that assesses standard semantic segmentation metrics on a per-connected component basis. CC-Metrics provides complementary information to standard global metrics, making it a valuable tool for evaluating model performances in the identified ``detection via segmentation'' scenario. Our proximity-based segmentation to ground-truth matching focuses on domains where the real-world distance between neighboring pixels is constant, such as in MRI or CT scans. While ideal for these cases, CC-Metrics can also be applied to any domain with constant pixel distance such as satellite imagery or microscopy images. 

\section{Acknowledgements}
The present contribution is supported by the Helmholtz Association under the joint research school “HIDSS4Health – Helmholtz Information and Data Science School for Health and was supported by funding from the pilot program Core-Informatics of the Helmholtz Association (HGF). This work was performed on the HoreKa supercomputer funded by the Ministry of Science, Research and the Arts Baden-Württemberg and by the Federal Ministry of Education and Research.

\bibliography{aaai25}
\clearpage

\twocolumn[
\begin{center}
    \LARGE \textbf{Supplementary Material: Every Component Counts: Rethinking the Measure of Success for Medical Semantic Segmentation in Multi-Instance Segmentation Tasks}
    \vspace{1em} 
\end{center}
]

\section{Computation of Generalized Voronoi Diagrams}

In this section, we discuss the algorithm to compute the Generalized Voronoi Diagrams $R'_C$. The algorithm first computes connected components so that each connected component in the original mask $S$ is distinguishable. We utilize the implementation by Silversmith~\cite{silversmith2021cc3d} for this purpose. Subsequently, we compute the Euclidean distance transform for each component separately. We use the scipy implementation~\cite{virtanen2020scipy} for this computation. However, since the scipy implementation computes the distance transform of the pixels inside each component, and we are interested in the distances of the pixels of the background class, we invert the pixel values, setting everything except the current component to the foreground and the pixels within the current component to the background. As this step is specific to the library used, we omit it from the pseudo-algorithm~\ref{Alg: Compute GVD}.
 
We collect the individual distance transforms and stack them in order. To compute the generalized Voronoi Regions $R'$ for the entire image, we take the argument of the minimal distance across all distance transforms. This operation assigns each voxel to its closest component as measured in Euclidean distance, which matches the definition of the generalized Voronoi Diagram.

\begin{algorithm}
\caption{Compute Generalized Voronoi Diagrams}
\begin{algorithmic}
\Require Segmentation mask $S\in \{0,1\}^{h\times w \times d}$
\Ensure Generalized Voronoi Regions $R'_C \quad \forall C \in \mathbb{K}$

\State \textbf{Step 1: Compute Connected Components}
\State $\mathbb{K} \gets \text{LabelConnectedComponents}(S)$

\State \textbf{Step 2: Compute Euclidean Distance Transforms}
\State $cc\_dt \gets [\ ]$
\For{each connected component $C$ in $\mathbb{K}$}
    \State $D_C \gets \text{euclidean\_dist\_transform}
    (\text{background of } C)$
    \State $cc\_dt.\text{append}(D_C)$
\EndFor
\State \textbf{Step 3: Compute Voronoi Regions}
\State $R'_C \gets {\text{argmin}}(\text{stack}(cc\_dt))$

\end{algorithmic}
\label{Alg: Compute GVD}
\end{algorithm}

\section{Proof on the Uniqueness of Generalized Voronoi Regions}

We want to ensure that the same target segmentation mask $S$ always leads to the same Voronoi regions for a given segmentation mask. We verify this for our generalized Voronoi diagram as specified in Def. 4 in the main paper.

\[
\begin{aligned}[b]
R'_C = & \left\{ t \in \mathbb{T} \mid d'(t, J_{C_k}) < d'(t, J_{C_l}) \right. \\
      & \left. \forall C_k, C_l \in \mathbb{K}, \, C_k \neq C_l \right\}
\end{aligned}
\]

\begin{theorem}
The generalized Voronoi Diagram as stated in Def. 4 is a unique separation of a metric space.
\end{theorem}
\noindent \textit{Proof:} It is a well-known fact that a standard Voronoi diagram is a unique separation of a metric space such as $\mathbb{T}$. We consider the case that in this standard diagram, each site consists of exactly one location (voxel) in each connected component $C$. 

\noindent Without the loss of generality, we choose one of the connected components and add one location which is $26$-connected to the one, that is currently forming the site. This operation may have the effect that some critical voxel locations $K_{crit}$ which have previously been within different regions are now closer to the enlarged site. These voxels in $K_{crit}$ however would only be closer to the now enlarged site, not an arbitrary other site. This still forms a unique separation of the metric space.

In the context of a discrete metric space, such as $\mathbb{T}$, the situation of boundary points is handled explicitly by the discrete nature of the space. Here, each point (voxel) is either part of a specific region or, in rare cases where distances are exactly equal, part of a shared boundary. In such cases, a consistent rule for boundary assignment is applied, ensuring the uniqueness of the separation. For example, boundary voxels may be assigned to the region associated with the site having the smallest index or based on a deterministic tie-breaking rule.

The described procedure of adding a location to a site can be repeated until all sites reflect $\mathbb{K}$ which fulfills Def. 4 of the Generalized Voronoi region, still forming a unique separation of $\mathbb{T}$. \qed

\section{Details on Synthethic Predictions and additional Simulation Results}
All synthetic predictions are simulated and evaluated on a Red Hat Linux machine with 152 cores and 256GB of RAM. To compute the used semantic segmentation metrics, their respective MONAI~\cite{cardoso2022monai} implementations are used. We compute CC-Metrics using Algorithm~\ref{Alg: Compute GVD} and run the same MONAI implementations per Voronoi Region $R'_C$. 

\subsubsection{Evaluation Results on a fixed data Subset}
\begin{figure*}
    \centering
    \setlength{\tabcolsep}{2pt}
    \begin{tabular}{>{\centering\arraybackslash}m{0.01\textwidth} 
                    >{\centering\arraybackslash}m{0.18\textwidth} 
                    >{\centering\arraybackslash}m{0.18\textwidth} |
                    >{\centering\arraybackslash}m{0.18\textwidth} 
                    >{\centering\arraybackslash}m{0.18\textwidth} 
                    >{\centering\arraybackslash}m{0.18\textwidth}}
        \multicolumn{3}{c|}{\textbf{Drop n Components}} & \multicolumn{3}{c}{\textbf{Increase Nof. Oversegmented Components}}\\
        
        & 
        \small Dice \& CC-Dice & \small Surface Dice \& CC-Surface Dice & \small Dice \& CC-Dice & \small Surface Dice \& CC-Surface Dice & \small Hausdorff95 Distance  \& CC-HD95 Distance \\
        
        \rotatebox{90}{\shortstack{\small \textbf{Edit n smallest}}} &
        \includegraphics[width=0.18\textwidth, trim=30 10 40 30, clip]{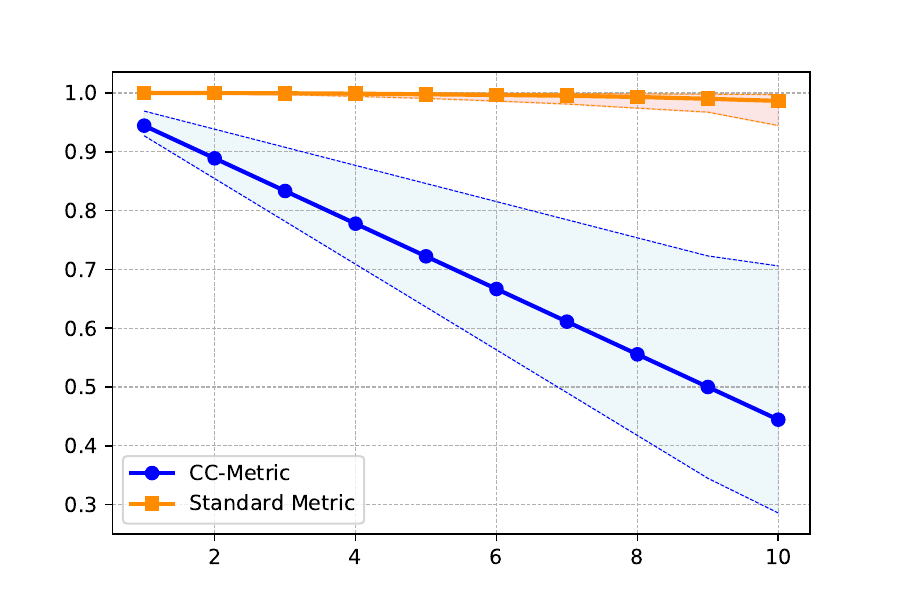} &
        \includegraphics[width=0.18\textwidth, trim=30 10 40 30, clip]{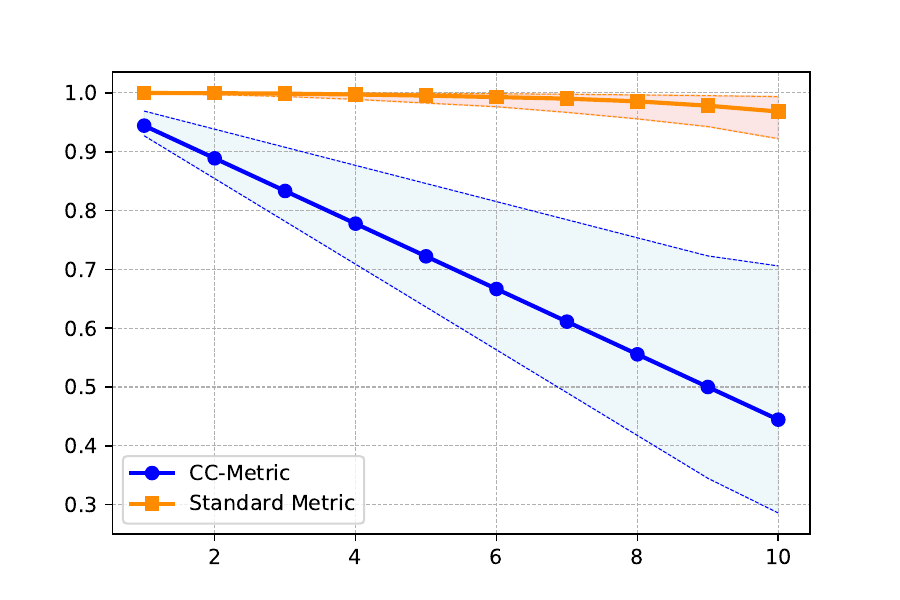} &
        \includegraphics[width=0.18\textwidth, trim=30 10 40 30, clip]{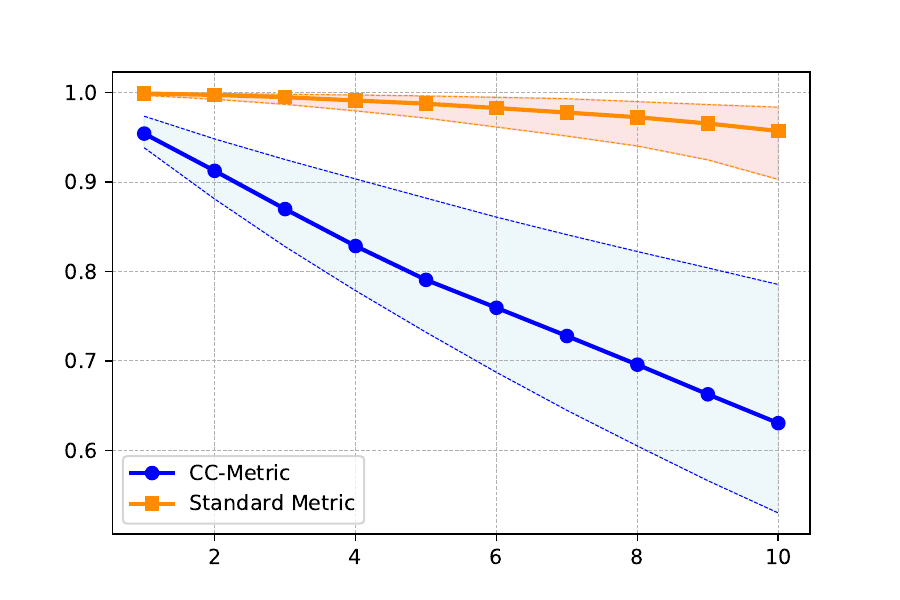} &
        \includegraphics[width=0.18\textwidth, trim=30 10 40 30, clip]{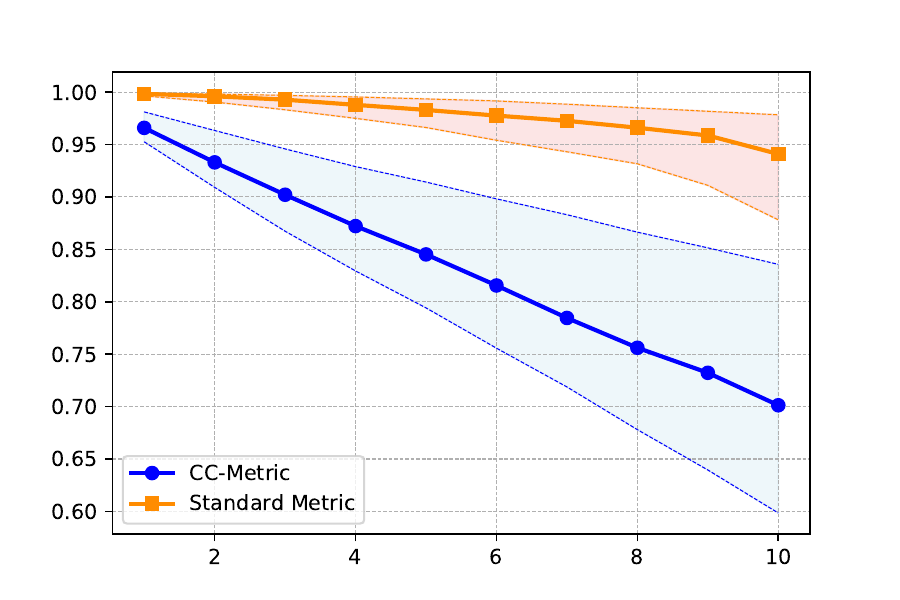} &
        \includegraphics[width=0.18\textwidth, trim=30 10 40 30, clip]{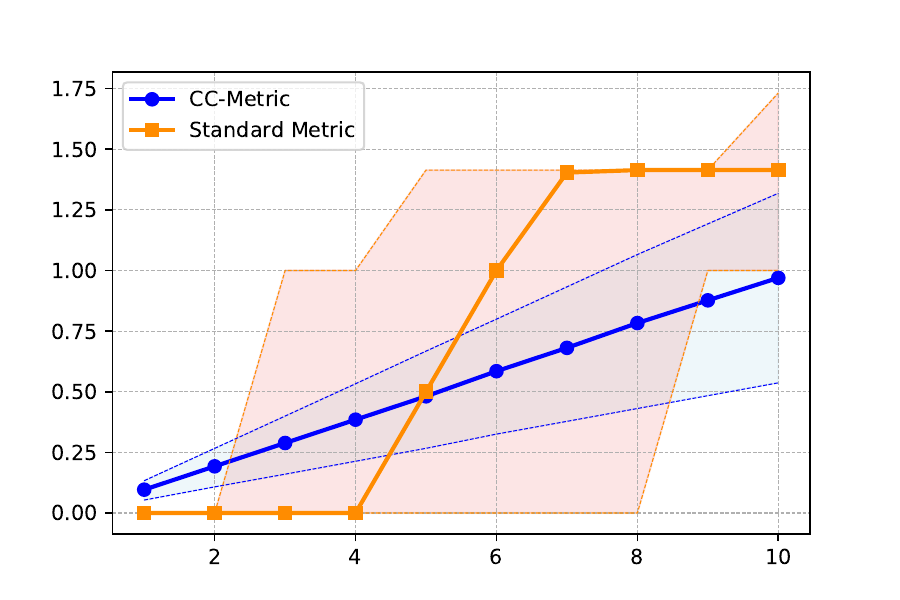} \\

        \rotatebox{90}{\shortstack{\small \textbf{Edit n largest}}} &
        \includegraphics[width=0.18\textwidth, trim=30 10 40 30, clip]{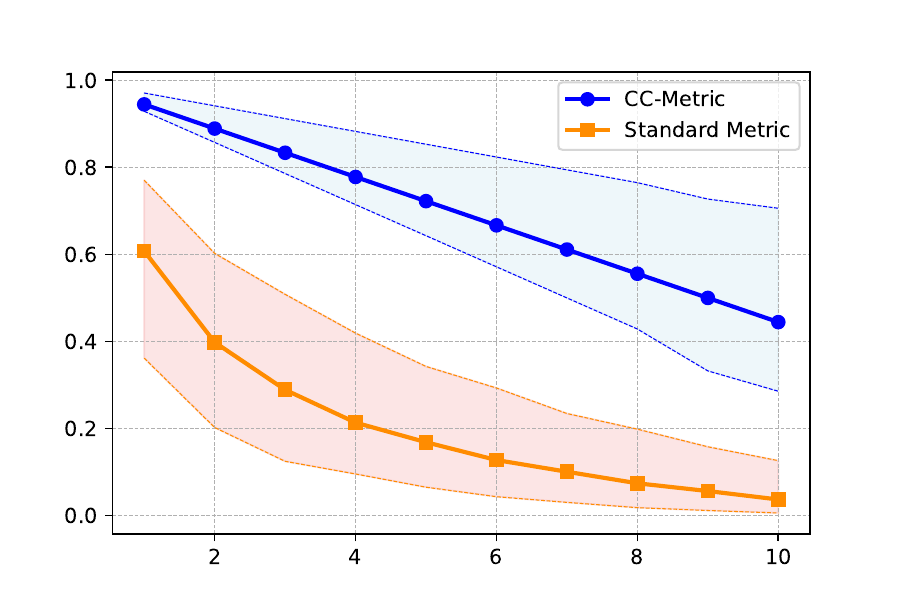} &
        \includegraphics[width=0.18\textwidth, trim=30 10 40 30, clip]{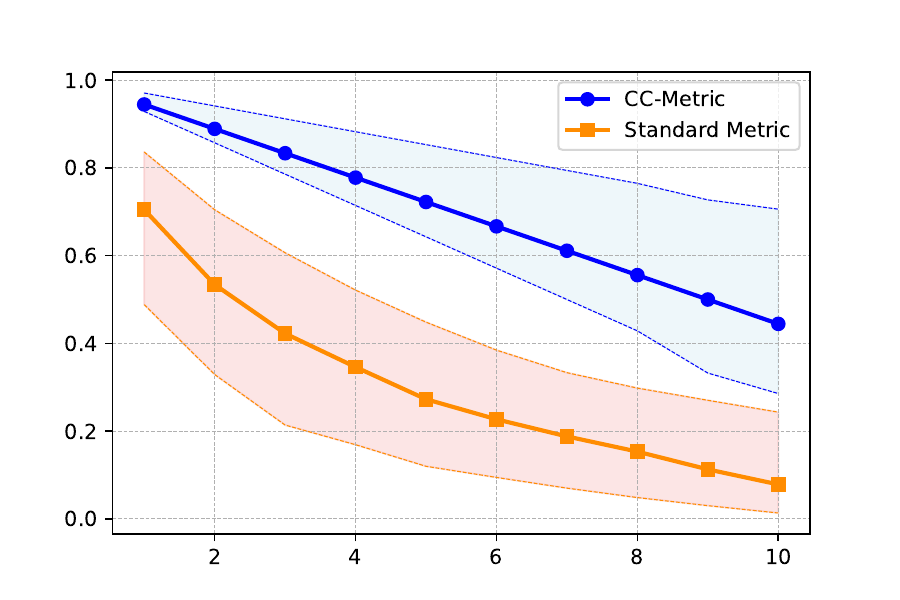} &
        \includegraphics[width=0.18\textwidth, trim=30 10 40 30, clip]{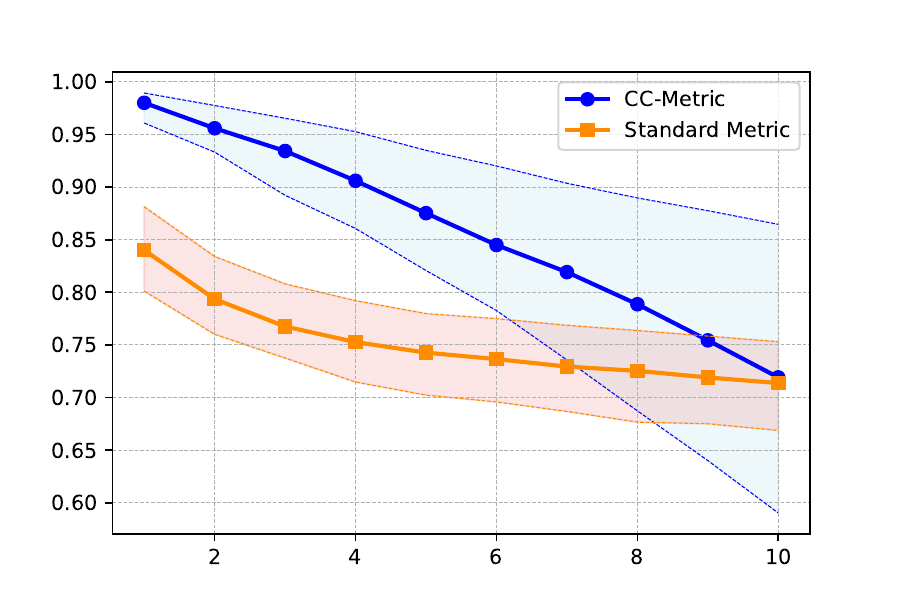} &
        \includegraphics[width=0.18\textwidth, trim=30 10 40 30, clip]{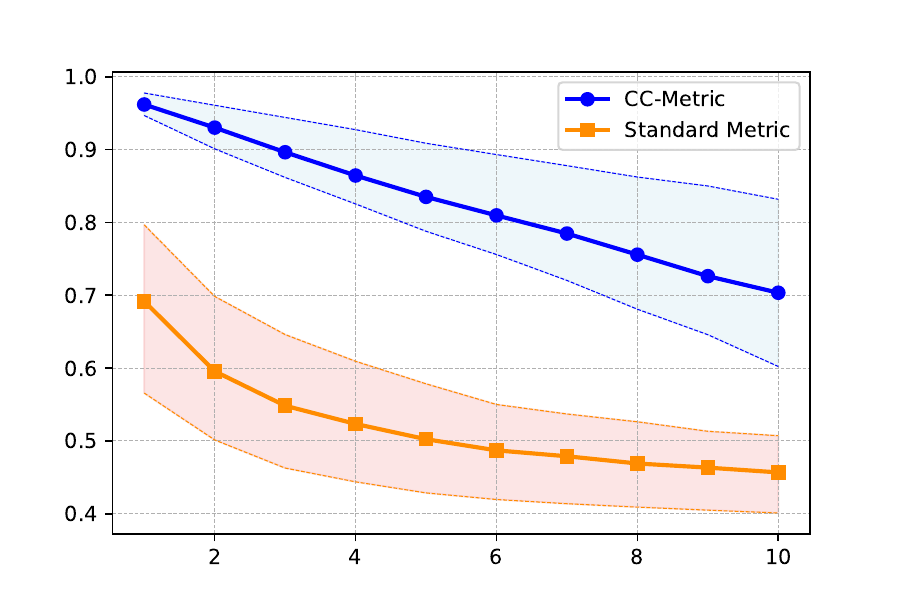} &
        \includegraphics[width=0.18\textwidth, trim=30 10 40 30, clip]{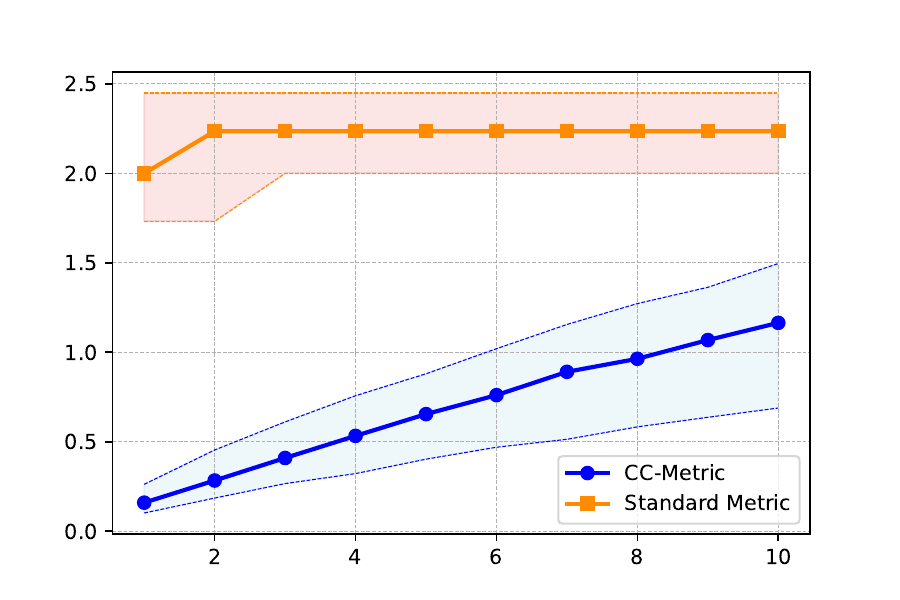} \\
    \end{tabular}


        \begin{tabular}{>{\centering\arraybackslash}m{0.01\textwidth} 
                    >{\centering\arraybackslash}m{0.18\textwidth} 
                    >{\centering\arraybackslash}m{0.18\textwidth} |
                    >{\centering\arraybackslash}m{0.18\textwidth} 
                    >{\centering\arraybackslash}m{0.18\textwidth} 
                    >{\centering\arraybackslash}m{0.18\textwidth}}
        \multicolumn{3}{c|}{\textbf{Insert n Random Components}} & \multicolumn{3}{c}{\textbf{Increase Nof. Undersegmented Components}}\\
        
        & 
        \small Dice \& CC-Dice & \small Surface Dice \& CC-Surface Dice & \small Dice \& CC-Dice & \small Surface Dice \& CC-Surface Dice & \small Hausdorff95 Distance  \& CC-HD95 Distance \\
        
        \rotatebox{90}{\shortstack{\small \textbf{Edit n smallest}}} &
        \includegraphics[width=0.18\textwidth, trim=30 10 40 30, clip]{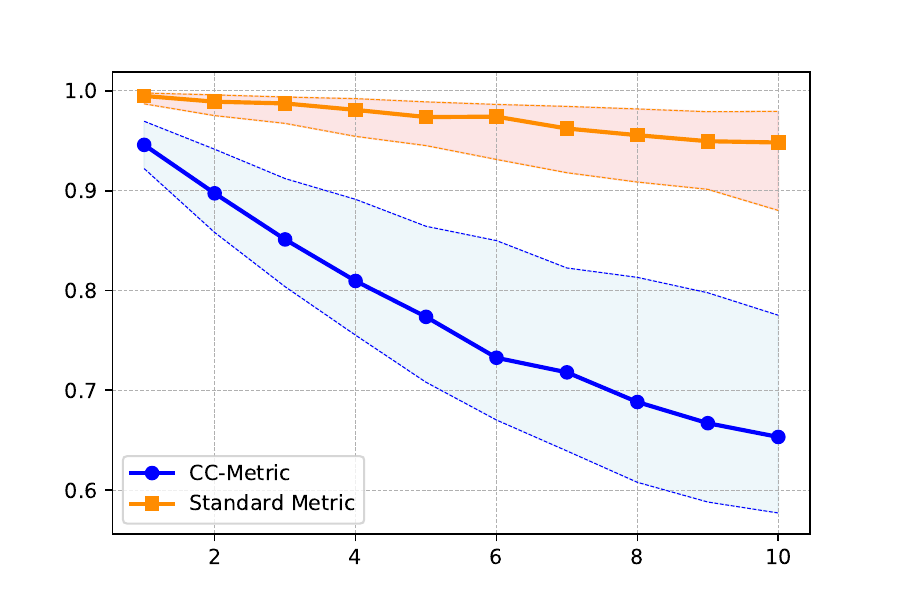} &
        \includegraphics[width=0.18\textwidth, trim=30 10 40 30, clip]{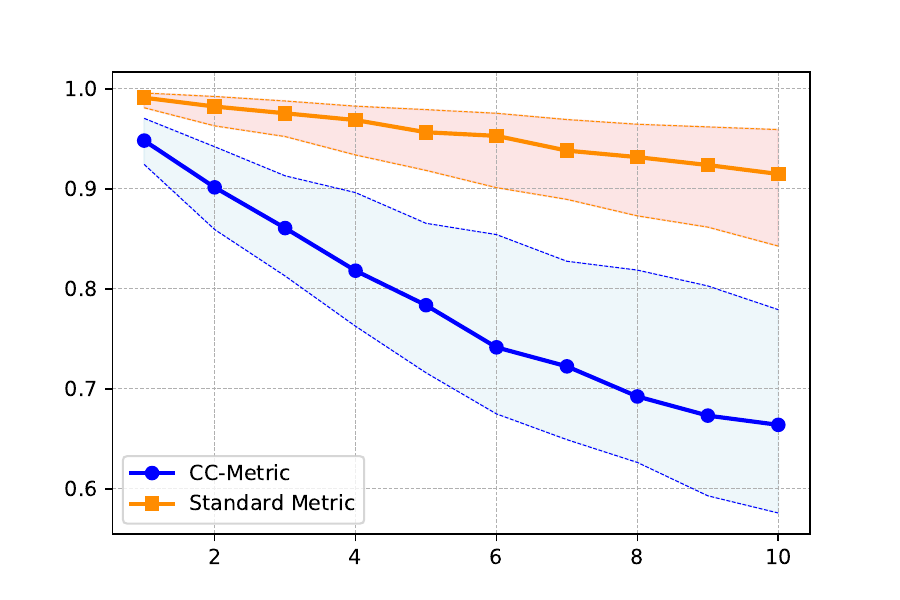} &
        \includegraphics[width=0.18\textwidth, trim=30 10 40 30, clip]{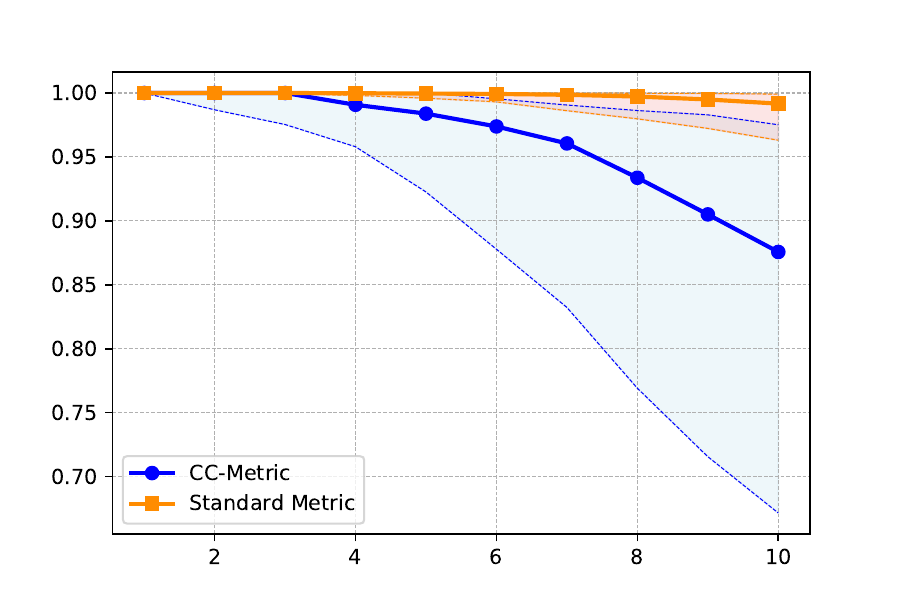} &
        \includegraphics[width=0.18\textwidth, trim=30 10 40 30, clip]{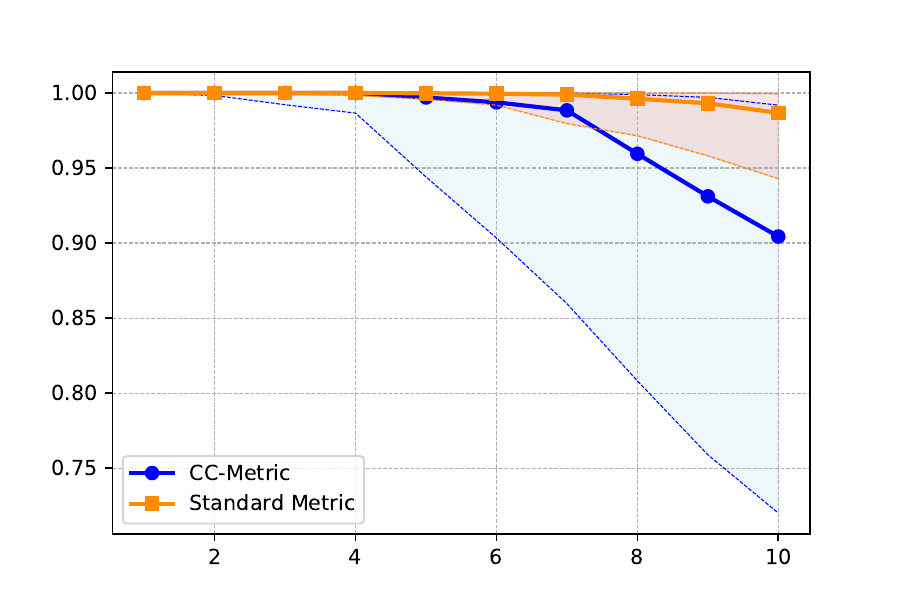} &
        \includegraphics[width=0.18\textwidth, trim=30 10 40 30, clip]{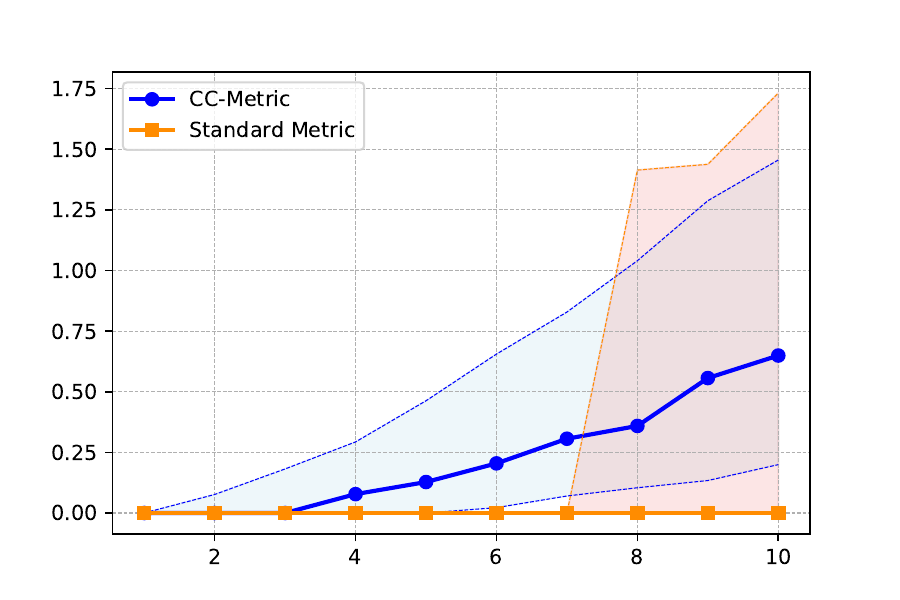} \\

        \rotatebox{90}{\shortstack{\small \textbf{Edit n largest}}} &
        \includegraphics[width=0.18\textwidth, trim=30 10 40 30, clip]{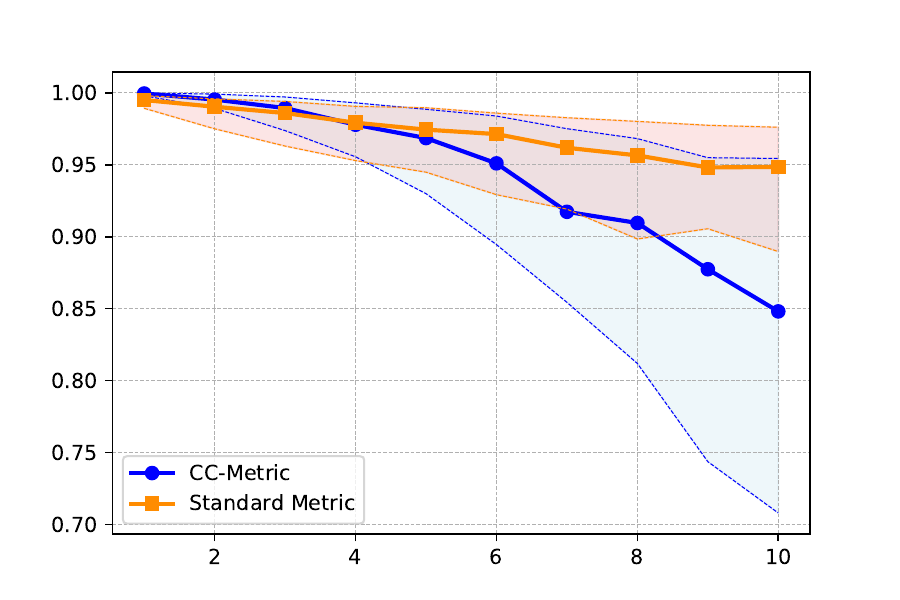} &
        \includegraphics[width=0.18\textwidth, trim=30 10 40 30, clip]{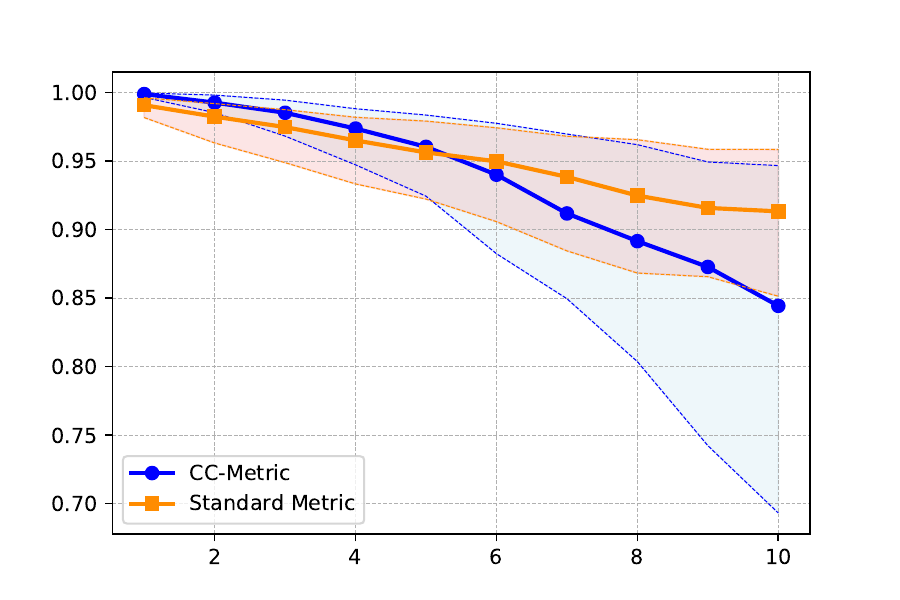} &
        \includegraphics[width=0.18\textwidth, trim=30 10 40 30, clip]{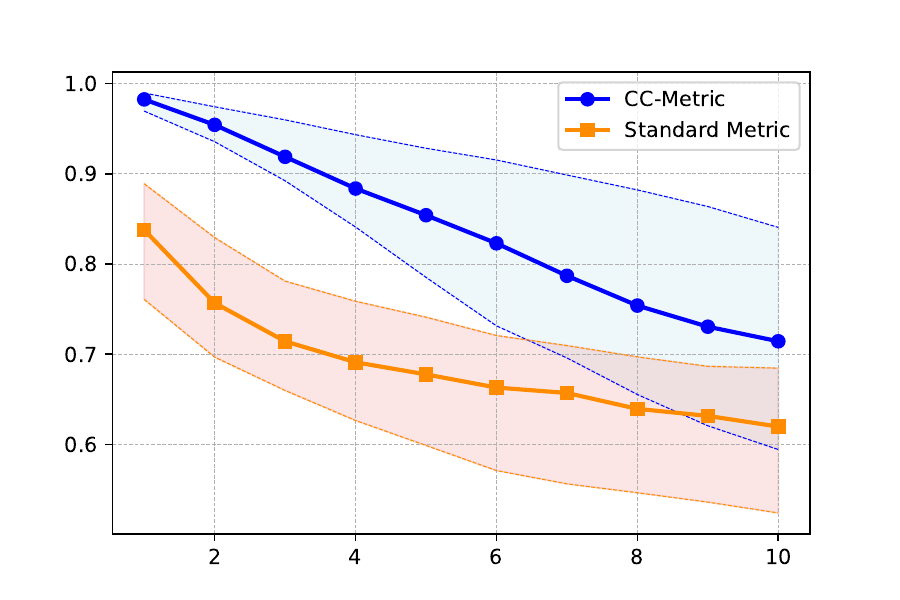} &
        \includegraphics[width=0.18\textwidth, trim=30 10 40 30, clip]{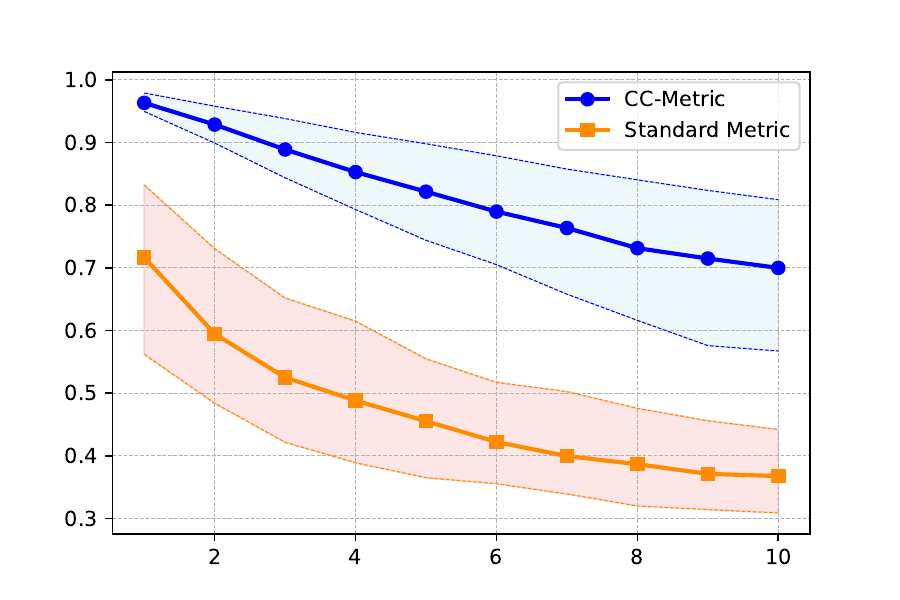} &
        \includegraphics[width=0.18\textwidth, trim=30 10 40 30, clip]{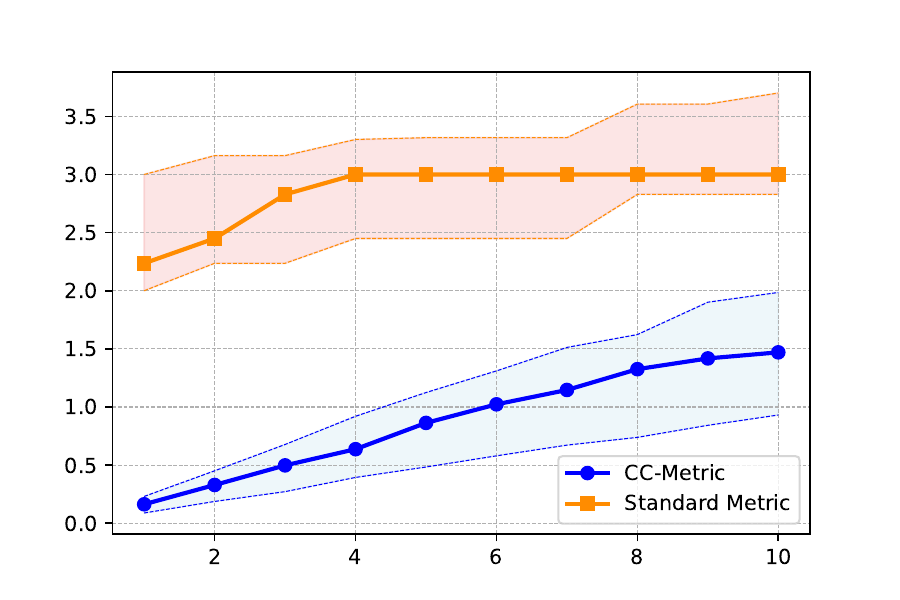} \\
    \end{tabular}
    
    \caption{Comparison of standard and CC semantic segmentation metrics on the AutoPET dataset across various scenarios. In this setting, we want to evaluate CC-Metrics on a constant subset of patients. Thus to drop a maximum of $n=10$ components, we include only patients with at least $11$ metastases for all measurements in this scenario including the ones where we drop less than $10$ components. For all other scenarios, we do not reduce the number of components during the prediction degradation and thus consider patients with at least $10$ components.
}
    \label{fig:autopet sym results appendix}
\end{figure*}

Fig.~4 in the original paper presents the evaluation results of synthetic predictions, where a degrading function progressively worsens an initially perfect prediction over $n$ editing steps. At each step, the analysis includes the maximum available data points, such as all patients with at least $n$ metastases when $n$ metastases are being edited. In the scenario where components are dropped, at least $n+1$ metastases are required when dropping $n$. While this approach maximizes the number of patients considered at each step, it also involves a different subset of patients at each step. Here, we present a complementary figure~\ref{fig:autopet sym results appendix}, where the subset of patients is kept constant by limiting the analysis to those with at least 10 metastases. For the ``Drop n Components'' scenario, we accordingly limit the analysis to patients with at least 11 metastases. The observations reported in the main paper remain valid in this scenario with a constant patient subset and are even more pronounced.

\section{Benchmark Model Training}
\label{APX: Benchmark Model Training}

\subsubsection{Datasets}
We base our experiments on the publicly available AutoPET-II~\cite{gatidis2023autopet} and HECKTOR~\cite{oreiller2022head} datasets to ensure a comprehensive analysis across different cancer types. The AutoPET-II dataset, which we also refer to as AutoPET for convenience, includes 1014 samples, consisting of patients diagnosed with malignant melanoma, lymphoma, or lung cancer, as well as negative control patients, whereas the HECKTOR dataset specifically targets head and neck tumors. Another distinguishing characteristic is the average number of tumors per patient in the two datasets: AutoPET features an average of more than 10 tumors per patient, in contrast to HECKTOR, which has only two tumors per patient on average. By utilizing both datasets, our experiments benefit from a diverse range of PET/CT data, providing a realistic assessment of the proposed CC-Metrics across scenarios with both high and low tumor counts. 

\subsubsection{Data Preprocessing}
As a first step, we align the CT and PET modalities in the HECKTOR dataset. To achieve this, we resample the PET to the CT resolution using third-order spline interpolation. This step is not necessary for AutoPET, as the authors release aligned images. As the normalization procedure, we take inspiration from the nnUNet~\cite{isensee2021nnu} framework and compute the fingerprints of the two datasets by computing the mean and the standard deviation of the CT and PET image values of the foreground regions. We employ a 5-fold cross-validation approach, where the dataset is divided into 5 folds. In each iteration, we train on 4 folds and predict on the remaining fold. This process is repeated 5 times to generate predictions for the entire dataset. To ensure representative folds, we use stratified sampling to generate the individual folds. For the AutoPET dataset, we maintain consistent distributions of cancer types and gender ratios across all folds. In the HECKTOR dataset, which lacks detailed cancer-type descriptions, we stratify only by gender. For all MONAI models, we exclude healthy patients to eliminate pure background samples, resulting in more informative gradients that accelerate the training process. This procedure is not required for nnU-Net, which is inherently trained for 1000 epochs.

\subsubsection{Implementation and Training Details}
We evaluate CC-Metrics along with standard evaluation protocols for four trained networks: nnUNet~\cite{isensee2021nnu}, DynUNet~\cite{isensee2019automated}, UNETR~\cite{hatamizadeh2022unetr}, and SwinUNETR~\cite{hatamizadeh2021swin}.
For the nnUNet, we leverage its out-of-the-box capabilities and do not change the default training setting. To handle both the CT and the PET input of the given PET/CT datasets, we concatenate the CT and the PET image as two channels and leave a third channel empty. 
For the training of DynUNet, UNETR, and SwinUNETR we use their respective MONAI~\cite{cardoso2022monai} implementations. 
We clip the CT image values between the $50^{\text{th}}$ and the $95^{\text{th}}$ percentile and the PET image values between the $1^{\text{st}}$ to $99.9^{\text{th}}$ percentile. We apply a Z-score normalization to each channel by subtracting the mean and dividing by the standard deviation, using the precomputed values from the entire CT and PET datasets.

The training for the Dynamic UNet on AutoPET was conducted over 200 epochs, with validations every 100 epochs. The training process utilized a sliding window approach with a patch size of (128, 128, 128) on AutoPET and (96, 96, 96) on HECKTOR as the images are spatially smaller. We crop 2 patches per image and use a batch size of 8. We use deep supervision and employ the AdamW optimizer, to update gradients produced by the DiceCELoss function. We set the initial learning rate to $10^{-3}$ and gradually reduce it using a Cosine Annealing LR scheduler.   
For the training of the transformer-based models, we use a patch size of (96, 96, 96) following the settings of the authors~\cite{hatamizadeh2021swin,hatamizadeh2022unetr}. We initialize the SwinUNETR model using the pre-trained weights provided by the authors. The transformer-based models are trained with an initial learning rate of $10^{-4}$ which is reduced using a Cosine Annealing scheduler. 

Except for the nnUNet, all models are trained with a batch size of 8 on 4 Nvidia A100 GPUs using a DDP-Setting, with 512GB of RAM on a Red Hat Linux machine with 152 cores. We parallelize the training using the PyTorch-Lightning framework~\cite{falcon2019pytorch}. 

\section{Qualitative Results}
\begin{figure*}
    \centering
    \includegraphics[width=0.75\linewidth, trim=0 30 340 0, clip]{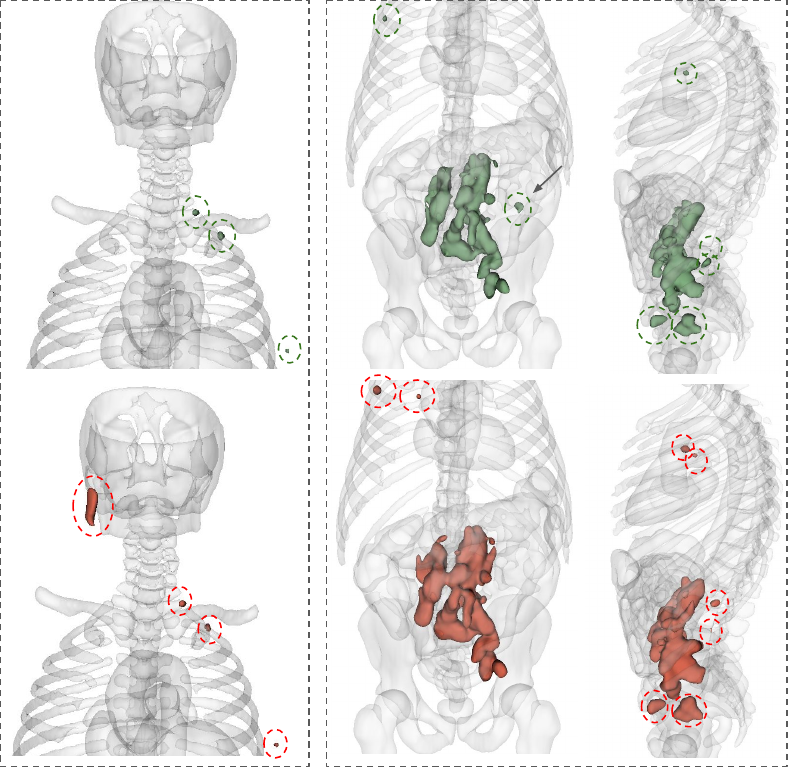}
    \caption{Qualitative Example of predictions (red) vs. ground truth (green) with large differences in standard Dice and CC-Dice. On the left, CC-Dice is higher than the standard Dice, while on the right, the standard Dice exceeds CC-Dice.}
    \label{fig: qualitative examples}
\end{figure*}

In Fig.~\ref{fig: qualitative examples} we show two examples of nnUNet predictions where standard Dice and CC-Dice disagree by a large margin. On the left, the Dice score of the prediction is $46.7\%$, while CC-Dice was reported as $74.4\%$. Prior to analyzing the qualitative example, one could hypothesize that the observed discrepancy might stem from the model's tendency to more accurately capture smaller predictions compared to larger ones. This is confirmed when observing the left plot, where the three small metastases are covered by the model, whereas the large metastasis in the right cheek is missed. This is however a rather rare example in the nnUNet predictions, as most of the patient-wise CC-Dice scores are lower than their standard Dice scores. A typical example where CC-Dice is much lower than standard Dice is shown on the right. The reported standard Dice for this example is $85.3\%$, whereas CC-Dice is only at $20.1\%$. While the predictions are well aligned, at first sight, there are subtle differences. For instance, a false positive is being segmented close to the aorta and a metastasis which is located on the right of the largest connected component (indicated by the arrow) has not been segmented by the network. Other metastases are either over- or undersegmented. While these subtle differences are not captured by the standard dice, some of these can have detrimental effects on the patient's projected survival time and may require a sudden change in the current cancer treatment plan. This example also gives an intuition on how the difference in standard metrics and CC-Metric reveals the patterns of errors in the network predictions. 
\end{document}